\documentclass[journal]{IEEEtran}

\usepackage[numbers,sort&compress]{natbib} 
\usepackage{float}

\usepackage{multirow}
\usepackage{threeparttable}

\usepackage[ruled,linesnumbered]{algorithm2e} 
\usepackage{amsmath}
\usepackage{amsfonts}
\usepackage{amssymb}  

\usepackage{graphicx}
\graphicspath{{images/}}                     

\usepackage{hyperref}

\usepackage[utf8]{inputenc}

\usepackage{pifont}

\usepackage{booktabs}

\usepackage{soul}
\usepackage{color, xcolor}
\soulregister\cite7
\soulregister\citep7
\soulregister\citet7 
\soulregister\ref7 
\soulregister\pageref7 

\begin{document}
%


\title{Remote Sensing Image Change Detection with Transformers}

\author{Hao Chen, Zipeng Qi and Zhenwei Shi$^\star$,  
\IEEEmembership{Member,~IEEE}

\thanks{
The work was supported by the National Key R\&D Program of China under the Grant 2019YFC1510905, the National Natural Science Foundation of China under the Grant 61671037 and the Beijing Natural Science Foundation under the Grant 4192034.
\emph{(Corresponding author: Zhenwei Shi (e-mail: shizhenwei@buaa.edu.cn))}}

\thanks{Hao Chen, Zipeng Qi and Zhenwei Shi are with the Image Processing Center, School of Astronautics, Beihang University, Beijing 100191, China, and with the Beijing Key Laboratory of Digital Media, Beihang University, Beijing 100191, China, and also with the State Key Laboratory of Virtual Reality Technology and Systems, School of Astronautics, Beihang University, Beijing 100191, China.}

}
\date{Jul. 2021}

\maketitle
\begin{abstract}
Modern change detection (CD) has achieved remarkable success by the powerful discriminative ability of deep convolutions. However, high-resolution remote sensing CD remains challenging due to the complexity of objects in the scene. Objects with the same semantic concept may show distinct spectral characteristics at different times and spatial locations. Most recent CD pipelines using pure convolutions are still struggling to relate long-range concepts in space-time. Non-local self-attention approaches show promising performance via modeling dense relations among pixels, yet are computationally inefficient. Here, we propose a bitemporal image transformer (BIT) to efficiently and effectively model contexts within the spatial-temporal domain. Our intuition is that the high-level concepts of the change of interest can be represented by a few visual words, i.e., semantic tokens. To achieve this, we express the bitemporal image into a few tokens, and use a transformer encoder to model contexts in the compact token-based space-time. The learned context-rich tokens are then feedback to the pixel-space for refining the original features via a transformer decoder. We incorporate BIT in a deep feature differencing-based CD framework. Extensive experiments on three CD datasets demonstrate the effectiveness and efficiency of the proposed method. Notably, our BIT-based model significantly outperforms the purely convolutional baseline using only 3 times lower computational costs and model parameters. Based on a naive backbone (ResNet18) without sophisticated structures (e.g., FPN, UNet), our model surpasses several state-of-the-art CD methods, including better than four recent attention-based methods in terms of efficiency and accuracy. Our code is available at https://github.com/justchenhao/BIT\_CD.
\end{abstract}


\IEEEpeerreviewmaketitle


\section{Introduction}
\label{sec:intro}

\IEEEPARstart{C}{hange} detection (CD) is one of the major topics in remote sensing (RS). The goal of CD is to assign binary labels (i.e., change or no change) to every pixel in a region by comparing co-registered images of the same region taken at different times \cite{SINGH1989}. The definition of change varies across applications, such as urban expansion \cite{Chen2020e}, deforestation \cite{Bem2020}, and damage assessment \cite{Xu2019}. Information extraction based on RS images still mainly relies on manual visual interpretation. Automatic CD technology can reduce abundant labor costs and time consumption, thus has raised increasing attention \cite{Shi2020, Chen2020, Chen2020e, Zhang2019c, Zhang2020a, Liu2019b, Zhang2020b, Peng2020a, Jiang2020, Diakogiannis2020}.

\begin{figure*}
        \centering
        \includegraphics[width=\textwidth]{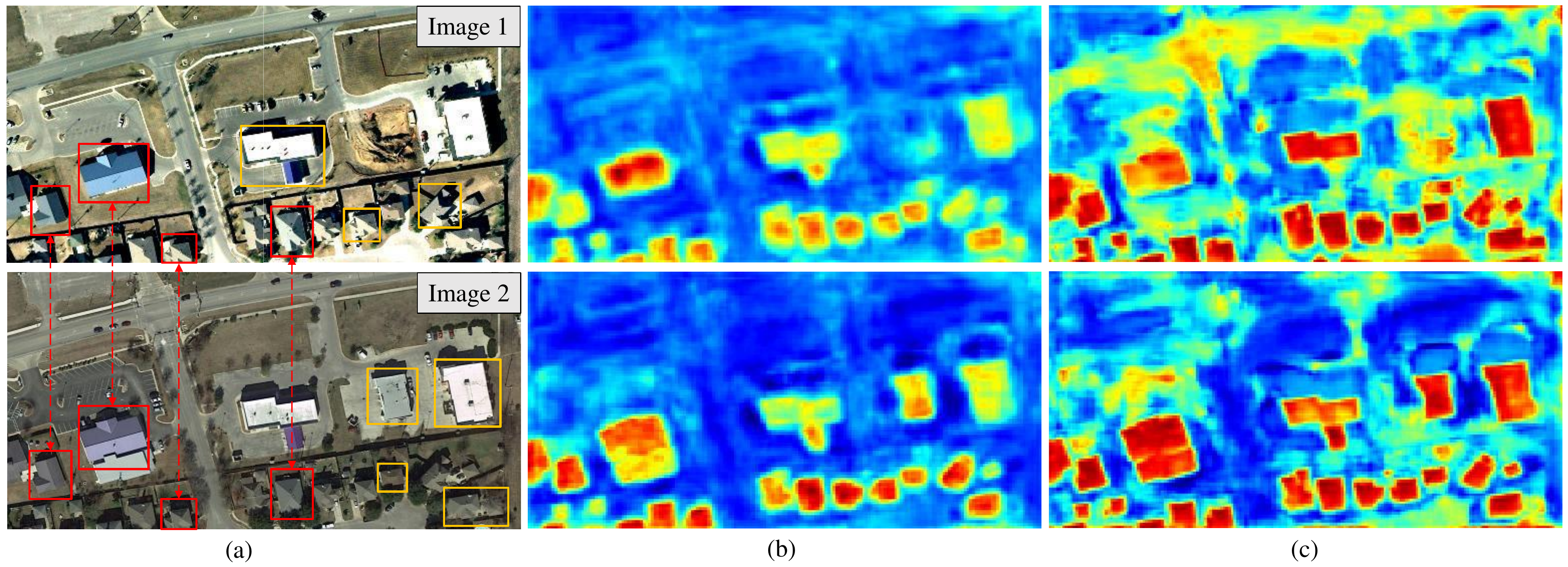}
        \caption{Illustration of the necessity of context modeling and the effect of our BIT module.
        (a) An example of a complex scene in bitemporal high-resolution images. Building objects show different spectral characteristics at different times (red boxes) and different spatial locations (yellow boxes). A strong building CD model needs to recognize the building objects and distinguish real changes from irrelevant changes by leveraging context information. Based on the high-level image features (b), our BIT module exploits global contexts in space-time to enhance the original features. The differencing image (c) between the enhanced features and the original one shows the consistent improvement in features of building areas across space-time.}
        \label{fig:image_appearance}
\end{figure*}

The availability of high-resolution (HR) satellite data and aerial data is opening up new avenues for monitoring land-cover and land-use at a fine scale. CD based on HR optical RS images remains a challenging task for two aspects: 1) complexity of the objects present in the scene, 2) different imaging conditions. Both contribute to the fact that the objects with the same semantic concept show distinct spectral characteristics at different times and different spatial locations (space-time). For example, as shown in Fig. \ref{fig:image_appearance} (a), the building objects in a scene have varying shapes and appearance (in yellow boxes), and the same building object at different times may have distinct colors (in red boxes) due to illumination variations and appearance alteration. To identify the change of interest in the complex scene, a strong CD model needs to, 1) recognize high-level semantic information of the change of interest in a scene, 2) distinguish the real change from the complex irrelevant changes.

Nowadays, due to its powerful discriminative ability, deep Convolutional Neural Networks (CNN) have been successfully applied in RS image analysis and have shown good performance in the CD task \cite{Shi2020}. Most recent supervised CD methods \cite{Chen2020, Chen2020e, Zhang2020a, Zhang2019c, Liu2019b, Zhang2020b, Peng2020a, Jiang2020, Diakogiannis2020} rely on a CNN-based structure to extract from each temporal image, high-level semantic features that reveal the change of interest. 

Since context modeling within the spatial and temporal scope is critical to identify the change of interest in high-resolution remote sensing images, the latest efforts have been focusing on increasing the reception field (RF) of the model, through stacking more convolution layers \cite{Chen2020, Chen2020e, Zhang2019c, Zhang2020a}, using dilated convolution \cite{Zhang2019c}, and applying attention mechanisms \cite{Liu2019b, Zhang2020b, Peng2020a, Jiang2020, Chen2020e, Chen2020, Diakogiannis2020}. Different from the purely convolution-based approach that is inherently limited to the size of the RF, the attention-based approach (channel attention \cite{Liu2019b, Zhang2020b, Peng2020a, Jiang2020}, spatial attention \cite{Liu2019b, Zhang2020b, Peng2020a}, and self-attention \cite{Chen2020e, Chen2020, Diakogiannis2020}) is effective in modeling global information. However, most existing methods are still struggling to relate long-range concepts in space-time, because they either apply attention separately to each temporal image for enhancing its features \cite{Liu2019b}, or simply use attention to re-weight the fused bitemporal features/images in the channel or spatial dimension \cite{Zhang2020b, Peng2020a, Jiang2020, Fang2021}. Some recent work \cite{Chen2020e, Chen2020, Diakogiannis2020} has achieved promising performance by utilizing self-attention to model the semantic relations between any pairs of pixels in space-time. However, they are computationally inefficient and need high computational complexity that grows quadratically with the number of pixels.

To tackle the above challenge, in this work, we introduce the Bitemporal Image Transformer (BIT) to model long-range context within the bitemporal image in an efficient and effective manner. Our intuition is that the high-level concepts of the change of interest could be represented by a few visual words, i.e., semantic tokens. Instead of modeling dense relations among pixels in pixel-space, our BIT expresses the input images into a few high-level semantic tokens, and models the context in a compact token-based space-time. Moreover, we enhance the feature representation of the original pixel-space by leveraging relations between each pixel and semantic tokens. Fig \ref{fig:image_appearance} gives an example to show the effect of our BIT on image features. Given the original image features related to the building concept (see Fig \ref{fig:image_appearance} (b)), our BIT learns to further consistently highlight the building areas (see Fig \ref{fig:image_appearance} (c)) by considering the global contexts in space-time. Note that we show the differencing image between the enhanced features and the original features to better demonstrate the role of the proposed BIT.

\begin{figure*}
        \centering
        \includegraphics[width=1\textwidth]{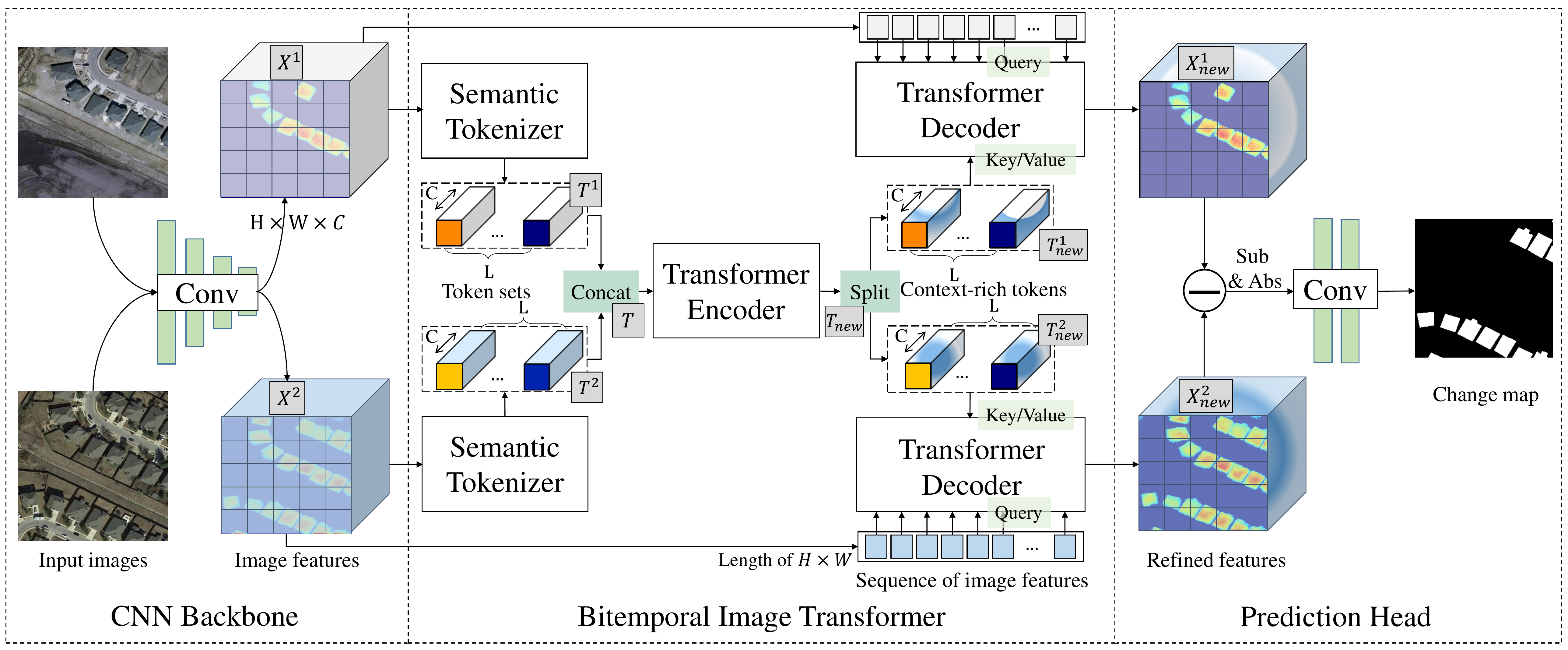}   
        \caption{Illustration of our BIT-based model. Our semantic tokenizer pools the image features extracted by a CNN backbone to a compact vocabulary set of tokens ($L<<HW$). Then we feed the concatenated bitemporal tokens to the transformer encoder to relate concepts in token-based space-time. The resulting context-rich tokens for each temporal image are projected back to the pixel-space to refine the original features via the transformer decoder. Finally, our prediction head produces the pixel-level predictions by feeding the computed feature difference images to a shallow CNN.}
        \label{fig:bit}
\end{figure*}

We incorporate BIT in a deep feature differencing-based CD framework. The overall procedure of our BIT-based model is illustrated in Fig. \ref{fig:bit}. A CNN backbone (ResNet) is used to extract high-level semantic features from the input image pair. We employ spatial attention to convert each temporal feature map into a compact set of semantic tokens. Then we use a transformer \cite{Vaswani2017} encoder to model the context within the two token sets. The resulting context-rich tokens are re-projected to the pixel-space by a Siamese transformer decoder for enhancing the original pixel-level features. Finally, we compute the Feature Difference Images (FDI) from the two refined feature maps, and then fed them into a shallow CNN to produce pixel-level change predictions.

The contribution of our work can be summarised as follows:
\begin{itemize}
\item An efficient transformer-based method is proposed for remote sensing image change detection. We introduce transformers into the CD task to better model contexts within the bitemporal image, which benefits to identify the change of interest and exclude irrelevant changes.

\item Instead of modeling dense relations among any pairs of elements in pixel-space, our BIT expresses the input images into a few visual words, i.e., tokens, and models the context in the compact token-based space-time.

\item Extensive experiments on three CD datasets validate the effectiveness and efficiency of the proposed method. We replace the last convolutional stage of ResNet18 with BIT, and the resulting BIT-based model outperforms the purely convolutional counterpart with a significant margin using only 3 times lower computational costs and model parameters. Based on a naive CNN backbone without sophisticated structures (e.g., FPN, UNet), ours shows better performance in terms of efficiency and accuracy than several recent attention-based CD methods.
\end{itemize}

The rest of this paper is organized as follows. Section \ref{sec:relatedwork} describes the related work of deep learning-based CD methods and the recent transformer-based models in RS. Section \ref{sec:method} gives the details of our proposed method. Some experimental results are reported in section \ref{sec:experiment}. The discussion is given in section \ref{sec:discussion} and conclusion is drawn in section \ref{sec:conclusion}.

\section{Related Work} \label{relatedwork}
\label{sec:relatedwork}

\subsection{Deep Learning based Remote Sensing Image Change detection}

Deep learning-based supervised CD methods for optical RS images can be generally divided into two main streams \cite{Zhang2020a}. 

One is the two-stage solution \cite{Nemoto2017, Ji2019, Liu2019a}, where a CNN/FCN is trained to separately classify the bitemporal images, and then their classification results are compared for change decision. This kind of approach is only practical when both the change label and the bitemporal semantic labels are available. 

Another is the single-stage solution, which directly produces the change result from the bitemporal images. The patch-level approach \cite{Daudt2018a, Rahman2018, Wang2020} models the CD task as a similarity detection process by grouping bitemporal images into pairs of patches and employing a CNN on each pair to obtain its center prediction. The pixel-level approach \cite{Daudt2018, Lebedev2018, Peng2019, Bem2020, Liu2019b, Jiang2020, Zhang2020b, Bao2020, Hou2020, Zhan2017, Fang2019, Zhang2019c, Chen2020, Chen2020e, Diakogiannis2020, Peng2020a} uses FCNs to directly generate a high-resolution change map from the two inputs, which is usually more efficient and effective than the patch-level approach. Since the CD task needs to handle two inputs, how to fuse the bitemporal information is an important topic. Existing FCN-based methods can be roughly divided into two groups according to the stage of fusion of bitemporal information. The image-level approach \cite{Daudt2018, Lebedev2018, Peng2019, Bem2020, Zhao2020c} concatenates the bitemporal images as a single input to a semantic segmentation network. The feature-level approach \cite{Daudt2018, Liu2019b, Jiang2020, Zhang2020b, Bao2020, Hou2020, Zhan2017, Fang2019, Zhang2019c, Chen2020, Chen2020e, Peng2020a, Chen2021a} combines the bitemporal features extracted from the neural networks and makes change decisions based on fused features. 

Much recent work aims to improve the feature discriminative power of the neural networks, by designing multi-level feature fusion structures \cite{Zhang2020b, Liu2019b, Jiang2020, Chen2020e, Hou2020, Chen2021a}, combining GAN-based optimization objectives \cite{Lebedev2018, Zhao2020b, Hou2020, Fang2019}, and increasing the reception field (RF) of the model for better context modeling in terms of the spatial and temporal scope \cite{Chen2020, Chen2020e, Zhang2019c, Zhang2020a, Liu2019b, Zhang2020b, Peng2020a, Jiang2020, Diakogiannis2020}. 

Context modeling is critical to identify the change of interest in high-resolution remote sensing images due to the complexity of the objects in a scene and the variation of image conditions. To increase the RF size, existing methods include employing a deeper CNN model \cite{Chen2020, Chen2020e, Zhang2019c, Zhang2020a}, using dilated convolution \cite{Zhang2019c}, and applying attention mechanisms \cite{Liu2019b, Zhang2020b, Peng2020a, Jiang2020, Chen2020e, Chen2020, Diakogiannis2020}. For example, Zhang et al. \cite{Zhang2019c} apply a deep CNN backbone (ResNet101 \cite{He2016}) to extract image features and use dilated convolution to enlarge the RF size of the model. Considering that purely convolutional networks are inherently limited to the size of the RF for each pixel, many latest efforts are focusing on introducing attention mechanisms to further enlarge the RF of the model, such as channel attention \cite{Liu2019b, Zhang2020b, Peng2020a, Jiang2020}, spatial attention \cite{Liu2019b, Zhang2020b, Peng2020a}, self-attention \cite{Chen2020e, Chen2020, Diakogiannis2020}. However, most of them still struggling to fully exploit the time-related context, because they either treat the attention as a feature enhancing module separately for each temporal image \cite{Liu2019b}, or simply use attention to re-weight the fused bitemporal features/images in the channel or spatial dimension \cite{Zhang2020b, Peng2020a, Jiang2020}. Non-local self-attention \cite{Chen2020e, Chen2020} shows promising performance due to its ability to exploit global relations among pixels in space-time. However, they are computationally inefficient and need high computational complexity that grows quadratically with the number of pixels. 

The main purpose of our paper is to learn and exploit the global semantic information within the bitemporal images in an efficient and effective manner for enhancing CD performance. Different from existing attention-based CD methods that directly model dense relations among any pairs of elements in pixel-based space, we extract a few semantic tokens from images and model the context in token-based space-time. The resulting context-rich tokens are then utilized to enhance the original features in pixel-space. Our intuition is that the change of interest within the scene can be described by a few visual words (tokens) and the high-level features of each pixel can be represented by the combination of these semantic tokens. As a result, our method exhibits high efficiency and high performance.

\subsection{Transformer-based Model}

The transformer, firstly introduced in 2017 \cite{Vaswani2017}, has been widely used in the field of natural language processing (NLP) to solve sequence-to-sequence tasks while handling long-range dependencies with ease. A recent trend is the adoption of transformers in the computer vision (CV) field. Due to the strong representation ability of the transformer, transformer-based models show comparable or even better performance as the convolutional counterparts in various visual tasks, including image classification \cite{Dosovitskiy2020, Touvron2020, Wu2020}, segmentation \cite{Zhang2020c, Wu2020, Zheng2020}, object detection \cite{Zhang2020c, Carion2020, Zhu2021}, image generation \cite{Chen2020h, Esser2020}, image captioning \cite{Liu2021}, and super-resolution \cite{Yang2020c, Chen2020g}.

The astounding performance of transformer models on NLP/CV tasks has intrigued the remote sensing community to study their applications in remote sensing tasks, such as image time-series classification \cite{Yuan2020, Li2020c}, hyperspectral image classification \cite{He2020a}, scene classification \cite{Bazi2021}, and remote sensing image captioning \cite{Shen2020a, Wang2020e}. For example, Li et al. \cite{Li2020c} proposed a CNN-transformer approach to perform the crop classification of time-series images, where the transformer was used to learn the pattern related to land cover semantics from the sequence of multitemporal features extracted via CNN. He et al. \cite{He2020a} applied a variant of the transformer (BERT \cite{Devlin2019}) to capture global dependencies among pixels in hyperspectral image classification. Moreover, Wang et al. \cite{Wang2020e} employed the transformer to translate the disordered words extracted by CNN from the given RS image into a well-formed sentence.

In this paper, we explore the potential of transformers in the binary CD task. Our proposed BIT-based method is efficient and effective in modeling global semantic relations in space-time to benefit the feature representation of the change of interest.

\section{Efficient Transformer based Change Detection Model}
\label{sec:method}

The overall procedure of our BIT-based model is illustrated in Fig. \ref{fig:bit}. We incorporate the BIT into a normal change detection pipeline because we want to leverage the strengths of both convolutions and transformers. Our model starts with several convolution blocks to obtain the feature map for each input image, then fed them into BIT to generate enhanced bitemporal features. Finally, the resulting feature maps are fed to a prediction head to produce pixel-level predictions. Our key insight is that BIT learns and relates the global context of high-level semantic concepts, and feedback to benefit the original bitemporal features.

Our BIT has three main components: 1) a Siamese semantic tokenizer, which groups pixels into concepts to generate a compact set of semantic tokens for each temporal input, 2) a transformer encoder, which models context of semantic concepts in token-based space-time, and 3) a Siamese transformer decoder, which projects the corresponding semantic tokens back to pixel-space to obtain the refined feature map for each temporal.

The inference detail of our BIT-based model for change detection is shown in Algorithm \ref{alg-inference}. 

\begin{algorithm}
\caption{Inference of BIT-based Model for Change Detection.} 
\label{alg-inference} 
\KwIn{$\mathbf{I}=\{(\mathbf{I}^{1},\mathbf{I}^{2})\}$ (a pair of registered images)}
\KwOut{$\mathbf{M}$ (a prediction change mask)}  
\BlankLine
    // step1: extract high-level features by a CNN backbone \\
    \For{$i$ in \{$1,2$\}}
    {
        $\mathbf{X}^{i}$ = CNN\_Backbone($\mathbf{I}^{i}$)  \\
    }
    // step2: use BIT to refine bitemporal image features \\
    // compute the token set for each temporal feature \\
    \For{$i$ in \{$1,2$\}}
    {
        $\mathbf{T}^{i}$ = Semantic\_Tokenizer($\mathbf{X}^{i}$)
    }
    $\mathbf{T}$=Concat($\mathbf{T}^{1}, \mathbf{T}^{2}$) \\
    // use encoder to generate context-rich tokens \\
    $\mathbf{T}_{new}$=Transformer\_Encoder($\mathbf{T}$) \\
    $\mathbf{T}^{1}_{new}, \mathbf{T}^{2}_{new}$=Split($\mathbf{T}_{new}$) \\
    // use decoder to refine the original features \\
    \For{$i$ in \{$1,2$\}}
    {
        $\mathbf{X}^{i}_{new}$ = Transformer\_Decoder($\mathbf{X}^{i}$, $\mathbf{T}^{i}_{new}$)
    }
    // step3: obtain change mask by the prediction head \\
    $\mathbf{M}$ = Prediction\_Head($\mathbf{X}^{1}_{new}, \mathbf{X}^{2}_{new}$)
   
\end{algorithm}

\begin{figure}
        \centering
        \includegraphics[width=0.5\textwidth]{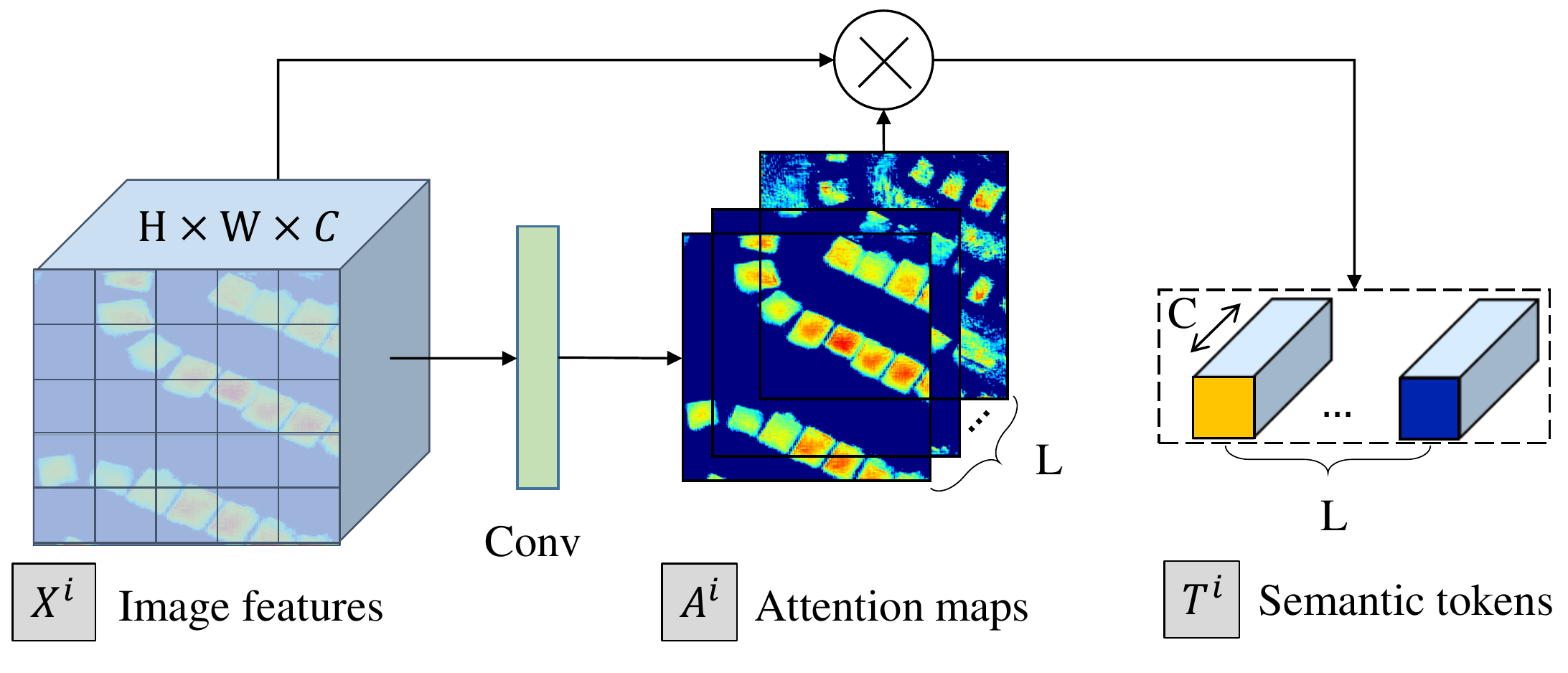}   
        \caption{Illustration of our semantic tokenizer.}
        \label{fig:tokenizer}
\end{figure}

\subsection{Semantic Tokenizer}
\label{ssec:tokenizer}

Our intuition is that the change of interest in input images could be described by a few high-level concepts, namely semantic tokens. And the semantic concepts can be shared by the bitemporal images. To this end, we employ a Siamese tokenizer to extract compact semantic tokens from the feature map of each temporal. Similar to the tokenizer in NLP, which splits the input sentence into several elements (i.e., word or phrase) and represents each element with a token vector, our semantic tokenizer splits the whole image into a few visual words, each corresponds to one token vector. As shown in Fig. \ref{fig:tokenizer}, to obtain the compact tokens, our tokenizer learns a set of spatial attention maps to spatially pool the feature map to a set of features, i.e., the token set. 

Let $\mathbf{X}^{1}, \mathbf{X}^{2} \in \mathbb{R}^{HW \times C}$ be the input bitemporal feature maps, where $H, W, C$ is height, width, and channel dimension of the feature map. Let $\mathbf{T}^{1}, \mathbf{T}^{2} \in \mathbb{R}^{L \times C}$ be the two sets of tokens, where $L$ is the size of the vocabulary set of tokens. 

For each pixel $\mathbf{X}^{i}_{p}$ on the feature map $\mathbf{X}^{i} (i=1,2)$, we use a point-wise convolution to obtain $L$ semantic groups, each group denotes one semantic concept. Then we compute spatial attention maps by a softmax function operated on the $HW$ dimension of each semantic group. Finally, we use the attention maps to compute the weighted average sum of pixels in $\mathbf{X}^{i}$ to obtain a compact vocabulary set of size $L$, i.e., semantic tokens $\mathbf{T}^{i}$. Formally, 
\begin{equation}
\mathbf{T}^{i} = (\mathbf{A}^{i})^{T}\mathbf{X}^{i}=(\sigma(\phi(\mathbf{X}^{i}; \mathbf{W})))^{T}\mathbf{X}^{i},
\end{equation}
where $\phi(\cdot)$ denotes the point-wise convolution with a learnable kernel $\mathbf{W} \in \mathbb{R}^{C \times L}$, $\sigma(\cdot)$ is the softmax function to normalize each semantic group to obtain the attention maps $\mathbf{A}^{i} \in \mathbb{R}^{HW \times L}$. $\mathbf{T}^{i}$ is computed by the multiplication of $\mathbf{A}^{i}$ and $\mathbf{X}^{i}$.

\begin{figure}
        \centering
        \includegraphics[width=0.5\textwidth]{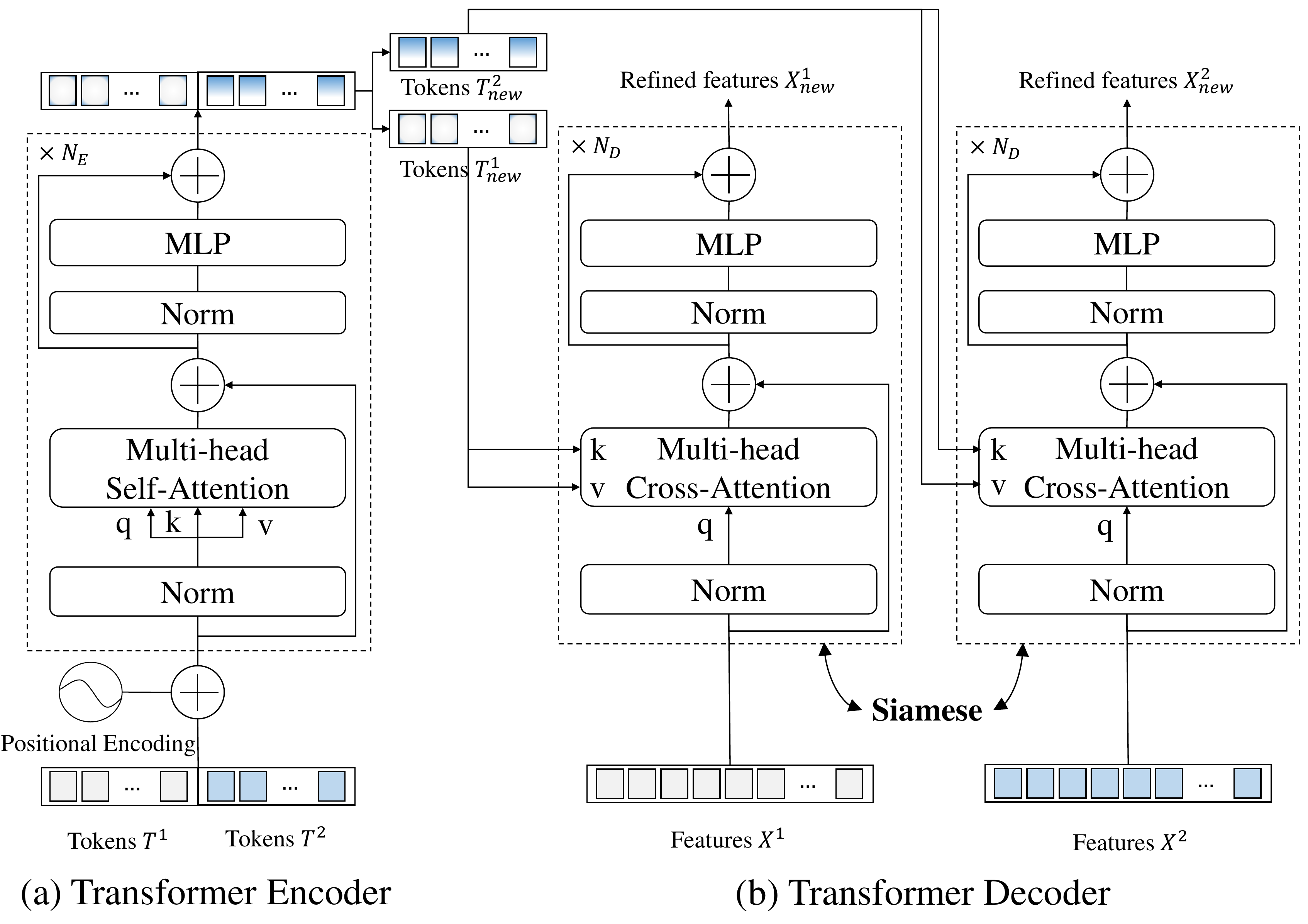}   
        \caption{Illustration of our transformer encoder and transformer decoder.}
        \label{fig:transformer}
\end{figure}

\subsection{Transformer Encoder}
\label{ssec:transformer_encoder}

After obtaining two semantic token sets $\mathbf{T}^{1},\mathbf{T}^{2}$ for the input bitemporal image, we then model the context between these tokens with a transformer encoder \cite{Vaswani2017}. Our motivation is that the global semantic relations in the token-based space-time can be fully exploited by the transformer, thus producing context-rich token representation for each temporal. As shown in Fig. \ref{fig:transformer} (a), we first concatenate the two sets of tokens into one token set $\mathbf{T} \in \mathbb{R}^{2L \times C}$, and fed it into the transformer encoder to obtain a new token set $\mathbf{T_{new}}$. Finally, we split the tokens into two sets $\mathbf{T}^{i}_{new} (i=1,2)$.

 The transformer encoder consists of $N_{E}$ layers of multi-head self-attention (MSA) and multilayer perceptron (MLP) blocks (Fig. \ref{fig:transformer} (a)). Different from the original transformer that uses the post-norm residual unit, we follow ViT \cite{Dosovitskiy2020} to adopt the pre-norm residual unit (PreNorm), i.e., the layer normalization occurs immediately before the MSA/MLP. PreNorm has been shown more stable and competent than the counterpart \cite{Nguyen2019}. 
 
 At each layer $l$, the input to self-attention is a triple (query $\mathbf{Q}$, key $\mathbf{K}$, value $\mathbf{V}$) computed from the input $\mathbf{T}^{(l-1)} \in \mathbb{R}^{2L \times C}$ as:
\begin{equation}
    \begin{split}
        \mathbf{Q} =  \mathbf{T}^{(l-1)} \mathbf{W}^{q},\\
        \mathbf{K} =  \mathbf{T}^{(l-1)} \mathbf{W}^{k},\\
        \mathbf{V} =  \mathbf{T}^{(l-1)} \mathbf{W}^{v},
    \end{split}
\end{equation}
 where $\mathbf{W}^{l-1}_{q}, \mathbf{W}^{l-1}_{k}, \mathbf{W}^{l-1}_{v} \in \mathbb{R}^{C \times d}$ are the learnable parameters of three linear projection layers and $d$ is the channel dimension of the triple. One attention head is formulated as:
 \begin{equation}
     \text{Att}(\mathbf{Q},\mathbf{K},\mathbf{V}) = \sigma \left(\frac{\mathbf{Q} \mathbf{K}^{T}}{\sqrt{d}}\right) \mathbf{V},
 \end{equation}
where $\sigma(\cdot)$ denotes the softmax function operated on the channel dimension. 

The core idea of the transformer encoder is multi-head self-attention. MSA performs multiple independent attention heads in parallel, and the outputs are concatenated and then projected to result in the final values. The advantage of MSA is that it can jointly attend to information from different representation subspaces at different positions. Formally, 
\begin{equation}
\begin{split}
    \text{MSA}(\mathbf{T}^{(l-1)}) = \text{Concat}(\text{head}_{1},..,\text{head}_{h})\mathbf{W}^{O},\\
    where \quad \text{head}_{j}=\text{Att}(\mathbf{T}^{(l-1)}\mathbf{W}^{q}_{j},\mathbf{T}^{(l-1)}\mathbf{W}^{k}_{j}, \mathbf{T}^{(l-1)}\mathbf{W}^{v}_{j}),
\end{split}
\end{equation}
where $\mathbf{W}^{q}_{j},\mathbf{W}^{k}_{j}, \mathbf{W}^{v}_{j} \in \mathbb{R}^{C \times d}, \mathbf{W}^{O} \in \mathbb{R}^{hd \times C}$ are the linear projection matrices, $h$ is the number of attention heads.

The MLP block consists of two linear transformation layers with a GELU \cite{Hendrycks2016} activation in between. The dimensionality of input and output is $C$, and the inner-layer has dimensionality $2C$. Formally, 

\begin{equation}
    \text{MLP}(\mathbf{T}^{(l-1)})=\text{GELU}(\mathbf{T}^{(l-1)}\mathbf{W}_{1})\mathbf{W}_{2}
\end{equation}
where $\mathbf{W}_{1} \in \mathbb{R}^{C \times 2C}, \mathbf{W}_{2} \in \mathbb{R}^{2C \times C}$ are the linear projection matrices.

Note that we add the learnable positional embedding (PE) $\mathbf{W}_{PE} \in \mathbb{R}^{2L\times C}$ to the token sequence $\mathbf{T}$ before feeding it to the transformer layers. Our empirical evidence (Sec. \ref{ssec:ablation}) indicates it is necessary to supplement PE to tokens. PE encodes the information about the relative or absolute position of elements in the token-based space-time. Such position information may benefit context modeling. For example, temporal positional information can guide transformers to exploit temporal-related contexts.

\subsection{Transformer Decoder}
\label{ssec:transformer_decoder}
Till now, we have obtained two sets of context-rich tokens $\mathbf{T}^{i}_{new} (i=1,2)$ for each temporal image. These context-rich tokens contain compact high-level semantic information that well reveals the change of interest. Now, we need to project the representation of concepts back to pixel-space to obtain pixel-level features. To achieve this, we use a modified Siamese transformer decoder \cite{Vaswani2017} to refine image features of each temporal. As shown in Fig. \ref{fig:transformer} (b), given a sequence of features $\mathbf{X}^{i}$, the transformer decoder exploits the relation between each pixel and the token set $\mathbf{T}^{i}_{new}$ to obtain refined features $\mathbf{X}^{i}_{new}$. We treat pixels in $\mathbf{X}^{i}$ as queries and tokens as keys. Our intuition is that each pixel can be represented by the combination of the compact semantic tokens.

Our transformer decoder consists of $N_{D}$ layers of multi-head cross attention (MA) and MLP blocks. Different from the original implementation in \cite{Vaswani2017}, we remove the MSA block to avoid abundant computation of dense relations among pixels in $\mathbf{X}^{i}$. We adopt PerNorm and the same configuration of MLP as the transformer encoder. In MSA, the query, key, and value are derived from the same input sequence, while in MA, the query is from the image features $\mathbf{X}^{i}$, and the key and value are from the tokens $\mathbf{T}^{i}_{new}$. Formally, at each layer $l$, MA is defined as:
\begin{equation}
\begin{split}
    \text{MA}(\mathbf{X}^{i,(l-1)},\mathbf{T}^{i}_{new})=\text{Concat}(\text{head}_{1},...,\text{head}_{h})\mathbf{W}^{O}, \\
    where \quad \text{head}_{j} = \text{Att}(\mathbf{X}^{i,(l-1)}\mathbf{W}^{q}_{j},\mathbf{T}^{i}_{new}\mathbf{W}^{k}_{j}, \mathbf{T}^{i}_{new}\mathbf{W}^{v}_{j}),\\
\end{split}
\end{equation}
where $\mathbf{W}^{q}_{j},\mathbf{W}^{k}_{j}, \mathbf{W}^{v}_{j} \in \mathbb{R}^{C \times d}, \mathbf{W}^{O} \in \mathbb{R}^{hd \times C}$ are the linear projection matrices, $h$ is the number of attention heads.

Note that we do not add PE to the input queries, because our empirical evidence (Sec. \ref{ssec:ablation}) shows no considerable gains when adding PE. 

\subsection{Network Details}
\label{ssec:network_details}

\textbf{CNN backbone}. We use a modified ResNet18 \cite{He2016} to extract bitemporal image feature maps. The original ResNet18 has 5 stages, each with downsampling by 2. We replace the stride of the last two stages to 1 and add a point-wise convolution (output channel $C=32$) behind ResNet to reduce the feature dimension, followed by a bilinear interpolation layer, thus obtaining the output feature maps with a downsampling factor of 4 to reduce the loss of spatial details. We name this backbone ResNet18\_S5. To validate the effectiveness of the proposed method, we also use two lighter backbone, namely ResNet18\_S4/ResNet18\_S3, which only uses the first four/three stages of the ResNet18. 

\textbf{Bitemporal image transformer}. According to parameter experiments in Sec. \ref{parameter_analysis}, we set token length $L=4$. We set the layer numbers of the transformer encoder to 1 and that of the transformer decoder to 8. The number of heads $h$ in MSA and MA is set to 8 and the channel dimension $d$ for each head is set to 8.

\textbf{Prediction head}. Benefiting from the high-level semantic features extracted by CNN backbone and BIT, a very shallow FCN is employed for change discrimination. Given two upsampled feature maps $\mathbf{X}^{1*}, \mathbf{X}^{2*} \in \mathbb{R}^{H_{0} \times W_{0} \times C}$ from the output of BIT ($H_{0}, W_{0}$ is the height, width of the original image, respectively), the prediction head is to generate the predicted change probability maps $P\in \mathbb{R}^{H_{0} \times W_{0} \times 2}$, which is given by 
\begin{equation}
P=\sigma(g(D))=\sigma(g(|X^{1*}-X^{2*}|)),
\end{equation}
where Feature Difference Images (FDI) $D\in \mathbb{R}^{H_{0} \times W_{0} \times C}$ is the element-wise absolute of the subtraction of the two feature maps, $g:\mathbb{R}^{H_{0} \times W_{0} \times C} \rightarrow \mathbb{R}^{H_{0} \times W_{0} \times 2}$ is the change classifier and $\sigma(\cdot)$ denotes a softmax function pixel-wisely operated on the channel dimension of the output of the classifier. The configuration of our change classifier is two $3 \times 3$ convolutional layers with BatchNorm. The output channel of each convolution is "32, 2".

In the inference phase, the prediction mask $M\in \mathbb{R}^{H_{0} \times W_{0}}$ is computed by a pixel-wise Argmax operation on the channel dimension of $P$.

\textbf{Loss function}. In the training stage, we minimize the cross-entropy loss to optimize the network parameters. Formally, the loss function is defined as:
\begin{equation}
    L=\frac{1}{H_{0}\times W_{0}} \sum_{h=1, w=1}^{H,W}l(P_{hw},Y_{hw}),
\end{equation}
where $l(P_{hw},y)=-log(P_{hwy})$ is the cross-entropy loss, and $Y_{hw}$ is the label for the pixel at location $(h,w)$.

\section{Experimental Results and Analysis} \label{sec:experiment}

\subsection{Experimental setup}
\label{ssec:setup}

We conduct experiments on three change detection datasets.

\textbf{LEVIR-CD} \cite{Chen2020e} is a public large scale building CD dataset. It contains 637 pairs of high-resolution (0.5m) RS images of size $1024\times 1024$. We follow its default dataset split (training/validation/test). For the limitation of GPU memory capacity, we cut images into small patches of size $256\times 256$ with no overlap. Therefore, we obtain 7120/1024/2048 pairs of patches for training/validation/test respectively.

\textbf{WHU-CD} \cite{Ji2019a} is a public building CD dataset. It contains one pair of high-resolution (0.075m) aerial images of size $32507\times 15354$. As no data split solution is provided in \cite{Ji2019a}, we crop the images into small patches of size $256 \times 256$ with no overlap and randomly split it into three parts: 6096/762/762 for training/validation/test respectively.

\textbf{DSIFN-CD} \cite{Zhang2020b} is a public binary CD dataset. It includes six large pairs of high-resolution (2m) satellite images from six major cities in China, respectively. The dataset contains the change of multiple kinds of land-cover objects, such as roads, buildings, croplands, and water bodies. We follow the default cropped samples of size $512\times 512$ provided by the authors. We have 3600/340/48 samples for training/validation/test respectively.

To validate the effectiveness of our BIT-based model, we set the following models for comparison:

\begin{itemize}
    \item \textbf{Base}: our baseline model that consists of the CNN backbone (ResNet18\_S5) and the prediction head.
    \item \textbf{BIT}: our BIT-based model with a light backbone (ResNet18\_S4).
\end{itemize}

To further evaluate the efficiency of the proposed method, we additionally set the following models:
\begin{itemize}
    \item \textbf{Base\_S4}: a light CNN backbone (ResNet18\_S4) + the prediction head.
    \item \textbf{Base\_S3}: a much light CNN backbone (ResNet18\_S3) + the prediction head.
    \item \textbf{BIT\_S3}: our BIT-based model with a much light backbone (ResNet18\_S3).
\end{itemize}

\textbf{Implementation details}. Our models are implemented on PyTorch and trained using a single NVIDIA Tesla V100 GPU. We apply normal data augmentation to the input image patches, including flip, rescale, crop, and gaussian blur. We use stochastic gradient descent (SGD) with momentum to optimize the model. We set the momentum to 0.99 and the weight decay to 0.0005. The learning rate is initially set to 0.01 and linearly decay to 0 until trained 200 epochs. Validation is performed after each training epoch, and the best model on the validation set is used for evaluation on the test set. 

\textbf{Evaluation Metrics}. We use the F1-score with regard to the change category as the main evaluation indices. F1-score is calculated by the precision and recall of the test as follows:
\begin{equation}
    F1 = \frac{2}{\text{recall}^{-1}+\text {precision}^{-1}},
\end{equation}

Additionally, precision, recall, Intersection over Union (IoU) of the change category, and overall accuracy (OA) are also reported. The above metrics are defined as follows: 
\begin{equation}
    \begin{split}
        \text{precision = TP / (TP + FP)}\\
        \text{recall = TP / (TP+FN)}\\
        \text{IoU = TP / (TP+FN+FP)} \\
        \text{OA = (TP+TN) / (TP+TN+FN+FP)} \\
    \end{split}
\end{equation}
where $\text{TP, FP, FN}$ represent the number of true positive, false positive, and false negative respectively.

\subsection{Comparison to state-of-the-art}
\label{ssec:comparison}

We make a comparison with several state-of-the-art methods, including three purely convolutional-based methods (FC-EF \cite{Daudt2018}, FC-Siam-Di \cite{Daudt2018}, FC-Siam-Conc \cite{Daudt2018}) and 
four attention-based methods (DTCDSCN \cite{Liu2019b}, STANet \cite{Chen2020e}, IFNet \cite{Zhang2020b} and SNUNet \cite{Fang2021}).

\begin{itemize}
    \item FC-EF \cite{Daudt2018}: Image-level fusion method, where the bitemporal images are concatenated as a single input to a fully convolutional network.
    \item FC-Siam-Di \cite{Daudt2018}: Feature-level fusion method, which employs a Siamese FCN to extract multi-level features and use feature difference to fuse the bitemporal information.
    \item FC-Siam-Conc \cite{Daudt2018}: Feature-level fusion method, which employs a Siamese FCN to extract multi-level features and use feature concatenation to fuse the bitemporal information.
    \item DTCDSCN \cite{Liu2019b}: Multi-scale feature concatenation method, which adds channel attention and spatial attention to a deep Siamese FCN, thus obtaining more discriminative features. Note that they also trained two additional semantic segmentation decoders under the supervision of the label maps of each temporal. We omit the semantic segmentation decoders for a fair comparison.
    \item STANet \cite{Chen2020e}: Metric-based Siamese FCN based method, which integrates the spatial-temporal attention mechanism to obtain more discriminative features.
    \item IFNet \cite{Zhang2020b}: Multi-scale feature concatenation method, which applies channel attention and spatial attention to the concatenated bitemporal features at each level of the decoder. Deep supervision (i.e., computing supervised loss at each level of the decoder) is used to better train the intermediate layers.
    \item SNUNet \cite{Fang2021}: Multi-scale feature concatenation method, which combines the Siamese network and NestedUNet\cite{Zhou2018} to extract high-resolution high-level features. Channel attention is applied to the features at each level of the decoder. Deep supervision is also employed to enhance the discrimination ability of intermediate features.
   
\end{itemize}

We implement the above CD networks using their public codes with default hyperparameters. 

\begin{figure*}
    \centering
    \includegraphics[width=1\textwidth]{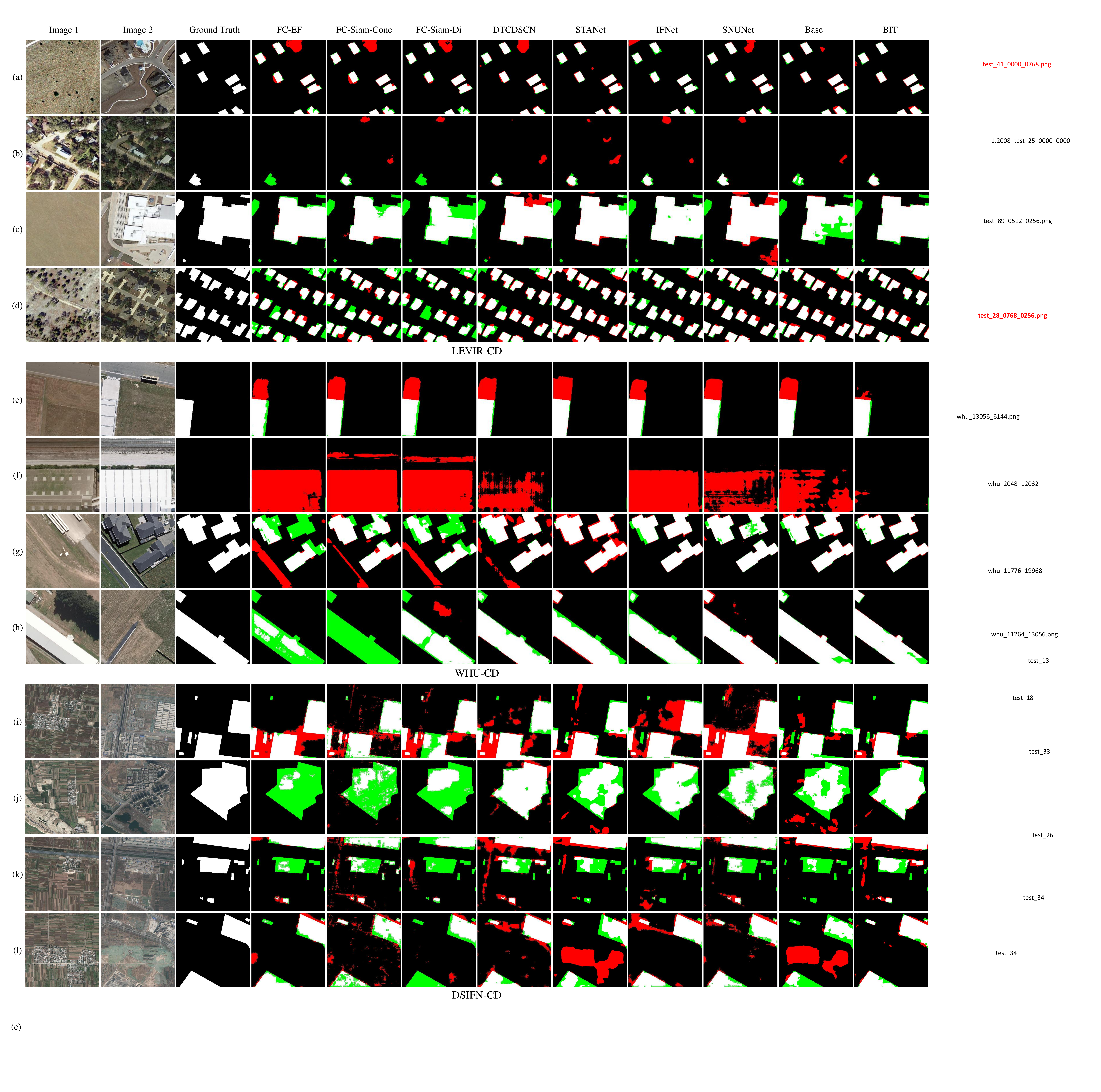}
    \caption{Visualization results of different methods on the LEVIR-CD, WHU-CD, and DSIFN-CD test sets. Different colors are used for a better view, i.e., white for true positive, black for true negative, red for false positive, and green for false negative.}
    \label{fig:visualization_all}
\end{figure*}

Tab. \ref{tab:comparison_sotas} reports the overall comparison results on LEVIR-CD, WHU-CD and DSIFN-CD test sets. The quantitative results show our BIT-based model consistently outperforms the other methods across these datasets with a significant margin. For example, the F1-score of our BIT exceeds the recent STANet by 2/1.6/4.7 points on the three datasets, respectively. Please note that our CNN backbone is only the pure ResNet and we do not apply the sophisticated structures such as FPN in \cite{Chen2020e} or UNet in \cite{Daudt2018, Liu2019b, Zhang2020b, Fang2021}, which are powerful for pixel-wise prediction tasks by fusing the low-level features with high spatial accuracy and high-level semantic features. We can conclude that even using a simple backbone, our BIT-based model can achieve superior performance. It may attribute to the ability of our BIT to model the context within the global highly abstract spatial-temporal scope and utilize the context for enhancing the feature representation in pixel-space.

The visualization comparison of the methods on the three datasets is displayed in Fig. \ref{fig:visualization_all}. For a better view, different colors are used to denote TP (white), TN (black), FP (red), FN (green). We can observe that the BIT-based model achieves better results than others. First, our BIT-based model can better avoid the false positive (e.g., Fig \ref{fig:visualization_all} (a), (e), (g), (i)) due to the similar appearance of the object as that of the interest change. For example, as shown in Fig. \ref{fig:visualization_all} (a), most comparison methods incorrectly classify the swimming pool area as the building change (view as red), while based on the enhanced discriminant features via global context modeling, the STANet and our BIT can reduce such false detection. In Fig. \ref{fig:visualization_all} (c), the roads are mistaken as building changes by conventional methods because the roads have similar color behaviors as buildings and these methods fail to exclude these pseudo changes due to their limited reception field. Second, our BIT can also well handle the irrelevant changes caused by seasonal differences or appearance alteration of land-cover elements (e.g., Fig \ref{fig:visualization_all} (b), (f) and (l)). An example of the non-semantic change of building in Fig \ref{fig:visualization_all} (f) illustrates the effectiveness of our BIT that learns the effective context within the spatial-temporal domain to better express the real semantic change and exclude the irrelevant change. Lastly, our BIT can generate relatively intact prediction results (e.g., Fig \ref{fig:visualization_all} (c), (h) and (j)) for large areas of change. For instance, in Fig. \ref{fig:visualization_all} (j), the large building area in image 2 can not be detected entirely (view as green) by some comparison methods due to their limited reception field, while our BIT-based model renders more complete results.

\begin{table*}
    \centering
    \caption{Comparison results on the three CD test sets. The highest score is marked in bold. All the scores are described in percentage (\%).}
    \resizebox{1\textwidth}{!}{
    \begin{tabular}{c|c|c|c}
  \toprule
    \multicolumn{1}{c}{} &
    \multicolumn{1}{|c|}{\textbf{LEVIR-CD}}  &  \multicolumn{1}{c|}{\textbf{WHU-CD}}  &
    \multicolumn{1}{c}{\textbf{DSIFN-CD}}  \\
    & Pre. / Rec. / F1 / IoU / OA & Pre. /  Rec. / F1 / IoU / OA & Pre. /  Rec. / F1 / IoU / OA \\
    \midrule
    FC-EF \cite{Daudt2018} & 86.91 / 80.17 / 83.40 / 71.53 / 98.39  &  71.63  /  67.25  /  69.37 / 53.11 / 97.61 & 72.61 / 52.73 / 61.09 / 43.98 / 88.59   \\
    FC-Siam-Di \cite{Daudt2018} & 89.53 / 83.31 /  86.31 / 75.92 / 98.67  &  47.33 / 77.66  /  58.81 / 41.66 /  95.63 & 59.67 / 65.71 / 62.54 / 45.50 / 86.63 \\
    FC-Siam-Conc \cite{Daudt2018} & \textbf{91.99}  / 76.77  / 83.69 / 71.96 / 98.49  & 60.88  /  73.58  /  66.63 / 49.95  / 97.04 & 66.45 / 54.21 / 59.71 / 42.56 / 87.57\\
    DTCDSCN \cite{Liu2019b} & 88.53  /  86.83  /  87.67 / 78.05 / 98.77 & 63.92  /  82.30  /  71.95 / 56.19 / 97.42 & 53.87 / \textbf{77.99} / 63.72 / 46.76 / 84.91 \\
    STANet \cite{Chen2020e} & 83.81  /  \textbf{91.00}  /  87.26 / 77.40 /  98.66 & 79.37  /  \textbf{85.50}  /  82.32 / 69.95 / 98.52 & 67.71 / 61.68 / 64.56 / 47.66 / 88.49  \\
    IFNet \cite{Zhang2020b} &  94.02 / 82.93 / 88.13 / 78.77 / 98.87 & 96.91 / 73.19 / 83.40 / 71.52 / 98.83 & 67.86 / 53.94 / 60.10 / 42.96 / 87.83 \\
    SNUNet \cite{Fang2021}  & 89.18 / 87.17 / 88.16 / 78.83 / 98.82 & 85.60 / 81.49 / 83.50 / 71.67 / 98.71 & 60.60 / 72.89 / 66.18 / 49.45 / 87.34 \\
    \midrule
    Base &  88.24 / 86.91 / 87.57 / 77.89 / 98.76 &  81.80 / 81.42 / 81.61 / 68.93 / 98.53 &  \textbf{73.30} / 48.65 / 58.48 / 41.32 / 88.26\\
    BIT & 89.24 / 89.37 / \textbf{89.31} / \textbf{80.68} / \textbf{98.92}  & \textbf{86.64} / 81.48 / \textbf{83.98} / \textbf{72.39} / \textbf{98.75}  & 68.36 / 70.18 / \textbf{69.26} / \textbf{52.97} / \textbf{89.41}\\
   \bottomrule
    \end{tabular}
    }
    \label{tab:comparison_sotas}
\end{table*}

\subsection{Model efficiency and effectiveness}

To fairly compare the model efficiency, we test all the methods on a computing server equipped with an Intel Xeon Silver 4214 CPU and an NVIDIA Tesla V100 GPU. Tab. \ref{tab:model_efficiency} reports the number of parameters (Params.), floating-point operations per second (FLOPs), and F1/IoU scores of different methods on LEVIR-CD, WHU-CD, and DSIFN-CD test sets. 

First, we verify the efficiency of our proposed BIT by comparing the convolutional counterparts. Tab. \ref{tab:model_efficiency} shows that built on Base\_S3/Base\_S4, the model added the BIT (BIT\_S3/BIT\_S4) is more effective and efficient than that (Base\_S4/Base\_S5) with more convolutional layers. For example, BIT\_S4 outperforms the Base\_S5 by 1.7/2.4/10.8 points of the F1-score on the three test sets while with 3 times smaller numbers of model parameters and 3 times lower computational costs. Moreover, we can observe that compared to Base\_S4, adding more convolutional layers only introduce trivial improvements (i.e., 0.16/0.75/0.18 points of the F1-score on the three test sets) while the improvements by BIT is much more (i.e., 4$\sim$60 times) than that of the CNN.

Second, we make a comparison with four attention-based methods (DTCDSCN, STANet, IFNet and SNUNet). As shown in Tab. \ref{tab:model_efficiency}, our BIT\_S4 outperforms the four counterparts in the F1/IoU scores with a significant margin with much small computational complexity and model parameters. Interestingly, even with a much lighter backbone (about 10 times smaller), our BIT-based model (BIT\_S3) is still superior to the four compared methods on most datasets. The comparison results further prove the effectiveness and efficiency of our BIT-based model.

\begin{table*}
    \centering
    \caption{Ablation study on model efficiency. We report the number of parameters (Params.), floating-point operations per second (FLOPs), as well as the F1 and IoU scores on the three CD test sets. The input image to the model has a resize of $256\times 256 \times 3$ to calculate the FLOPs.}
    \begin{tabular}{l|cc|cc|cc|cc}
    \toprule
    \multicolumn{3}{c}{} &
    \multicolumn{2}{|c|}{\textbf{LEVIR-CD}}  &  \multicolumn{2}{c|}{\textbf{WHU-CD}}  &
    \multicolumn{2}{c}{\textbf{DSIFN-CD}}  \\
    Model & Params.(M) & FLOPs (G) & F1  &  IoU  & F1  &  IoU  & F1  &  IoU  \\
    \midrule
        DTCDSCN \cite{Liu2019b} & 41.07 & 7.21 & 87.67 & 78.05 & 71.95 & 56.19 & 63.72 & 46.76 \\
        STANet \cite{Chen2020e} & 16.93 & 6.58 & 87.26 & 77.40  & 82.32 & 69.95 & 64.56 & 47.66 \\
        IFNet \cite{Zhang2020b} & 50.71 & 41.18 & 88.13 & 78.77  &  83.40 & 71.52  &  60.10 & 42.96  \\
        SNUNet \cite{Fang2021}  & 12.03 & 27.44 & 88.16  & 78.83 &  83.50 &  71.67 &   66.18  &  49.45 \\
        
    \midrule
        Base\_S3 & 1.28 & 1.78  & 82.23 & 76.24  & 79.52 & 66.00 & 56.00 & 38.88 \\
        + CNN (Base\_S4) & 3.38 & 4.09 & 87.41 & 77.64 & 80.86 & 67.87 & 58.30 & 41.15 \\
        + BIT (BIT\_S3)  & 1.45 & 2.05  & 88.51 & 79.39 & 81.38 & 68.60 &  69.00 & 52.67 \\
    \midrule
        Base\_S4 & 3.38 & 4.09 & 87.41 & 77.64 & 80.86 & 67.87 & 58.30 & 41.15 \\
        \multirow{1}{*}{+CNN (Base\_S5)}
         & 11.85 & 12.99 &  87.57&77.89  & 81.61 & 68.93 & 58.48 & 41.32\\
        \multirow{1}{*}{+BIT (BIT\_S4)}
        & 3.55  & 4.35 & \textbf{89.31} & \textbf{80.68}  & \textbf{83.98} & \textbf{72.39} & \textbf{69.26} & \textbf{52.97}\\
    \bottomrule
    \end{tabular}
    \label{tab:model_efficiency}
\end{table*}

\begin{figure}
\begin{minipage}[b]{1.0\linewidth}
  \centering
    \includegraphics[scale=0.48]{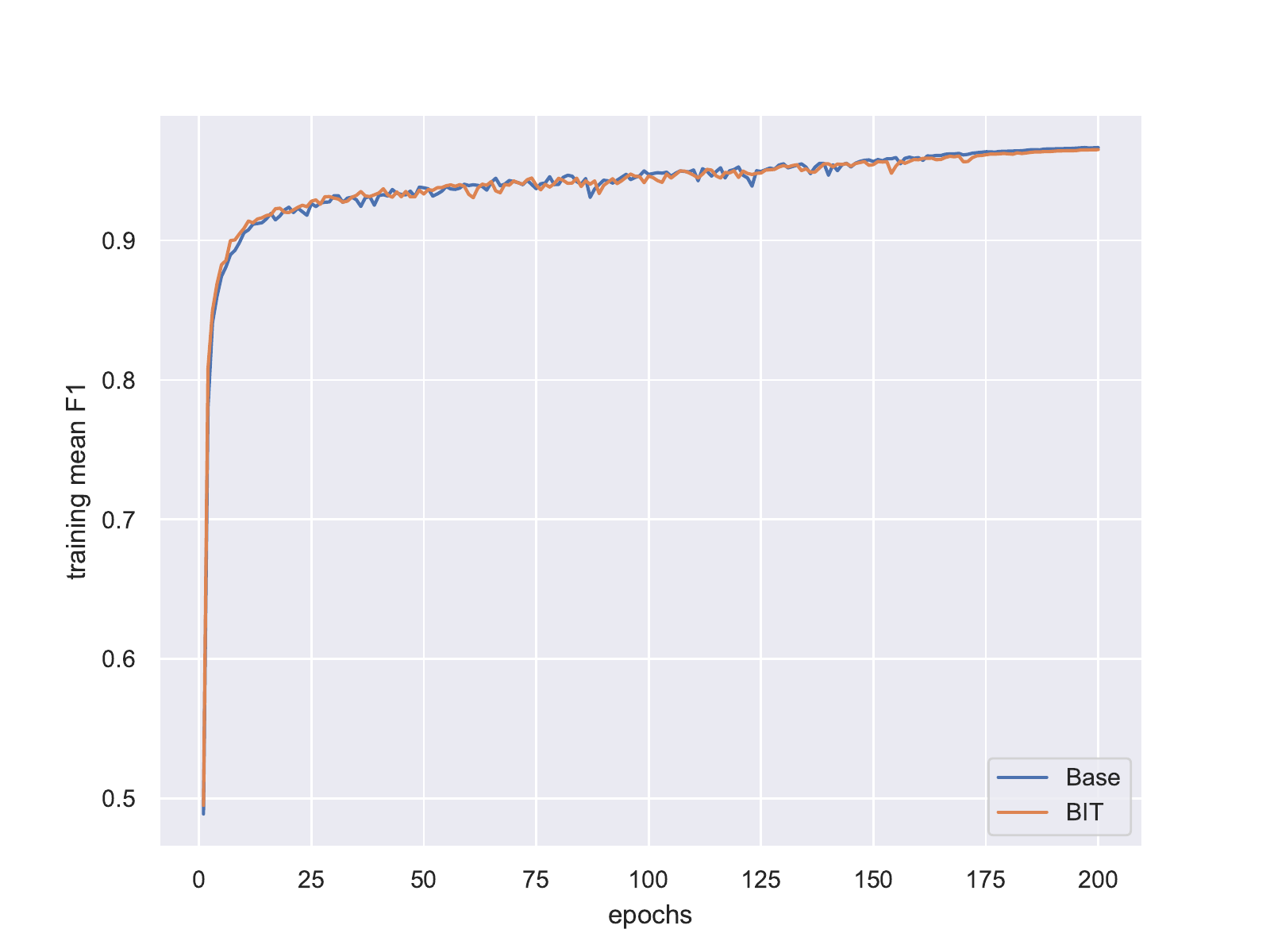} 
  \centerline{(a) Training accuracy on LEVIR-CD dataset.}\medskip
\end{minipage}
\hfill
\begin{minipage}[b]{1.0\linewidth}
  \centering
    \includegraphics[scale=0.48]{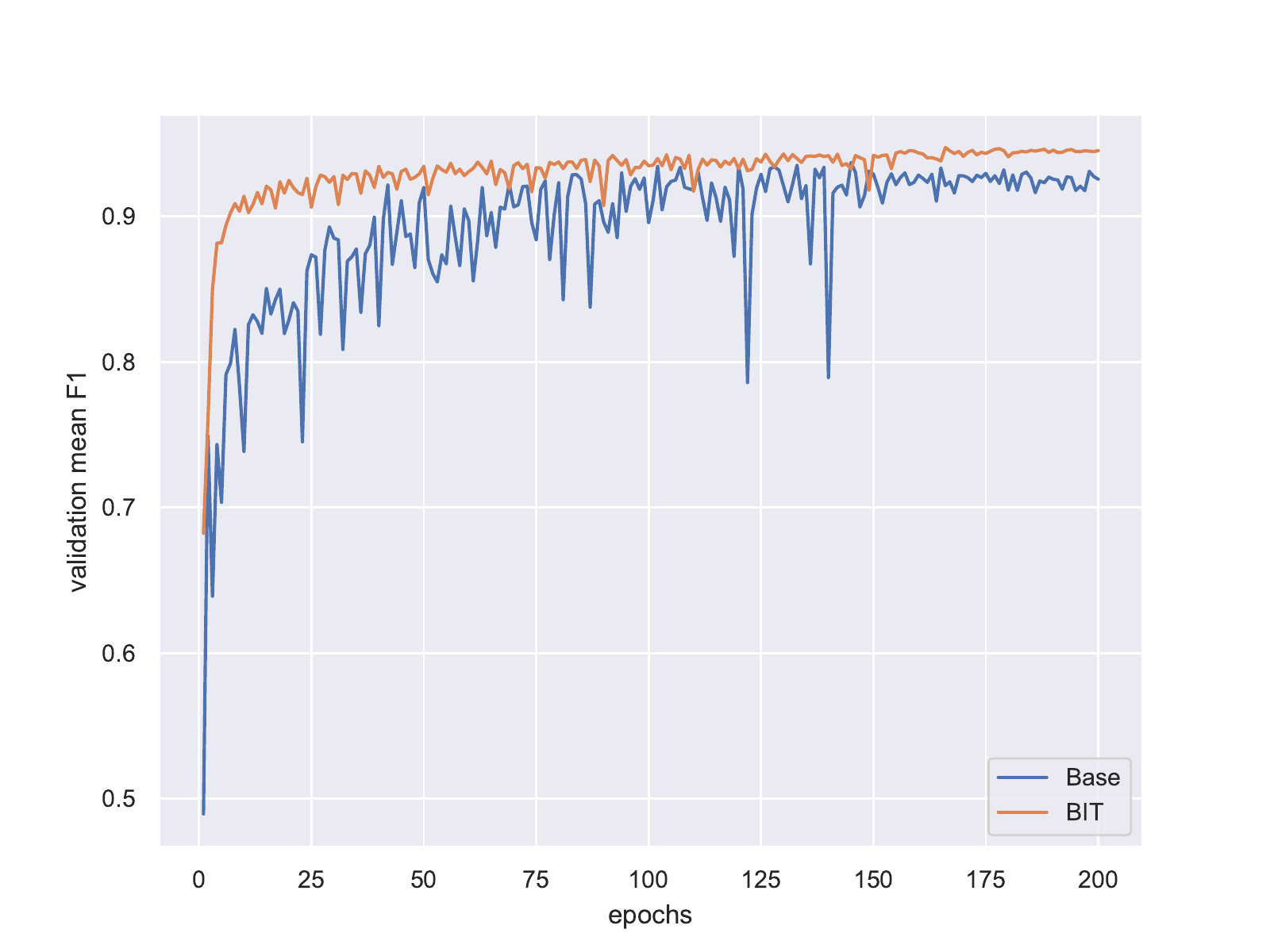} 
  \centerline{(b) Validation accuracy on LEVIR-CD dataset.}\medskip
\end{minipage}
\caption{Accuracy of models for each training epoch. The mean F1-score is reported.}
\label{fig:training_acc}
\end{figure}

\begin{table}
    \centering
    \caption{Ablation study of our BIT on three CD datasets. Ablations are performed on Tokenizer (T), Transformer Encoder (TE), and Transformer Decoder (TD). We also add the Non-local to the baseline for comparison. The F1-score is reported. Note that the depth of TE and TD are set to 1.}
    \begin{tabular}{ccccccc}
    \toprule
    Model & T & TE & TD & LEVIR & WHU & DSIFN \\
    \midrule
    Base\_S4 & $\times$ & $\times$ & $\times$  & 87.41 & 80.86 & 58.30  \\ 
    +Non-local & $\times$ & $\times$ & $\times$ & 87.56 & 80.93 & 59.94 \\
    \midrule 
    BIT & $\checkmark$ & $\checkmark$ & $\checkmark$ & \textbf{88.93} & \textbf{82.34} & \textbf{67.38} \\
    BIT & $\times$ & $\checkmark$ & $\checkmark$ & 87.58 & 81.68 & 61.76 \\
    BIT & $\checkmark$ & $\times$ & $\checkmark$  & 87.35 & 81.05 & 62.93 \\
    BIT & $\checkmark$ & $\checkmark$  & $\times$ & 88.07 & 79.16 & 64.47 \\
    BIT & $\checkmark$ & $\times$  & $\times$ & 87.38 & 80.82 & 59.54 \\
    \bottomrule
    \end{tabular}
    \label{tab:abalation_bit}
\end{table}

\begin{table}
    \centering
    \caption{Ablation study of position embedding (PE) on three CD datasets. We perform ablations on PE in Transformer Encoder (TE) and Transformer Decoder (TD). The F1-score is reported. Note that the depth of TE and TD are set to 1.}
    \begin{tabular}{cccccc}
    \toprule
    Model  &  PE in TE &  PE in TD & LEVIR & WHU & DSIFN \\
    \midrule
    BIT  & $\times$ & $\times$ & 87.77 & 82.06  & 60.81\\
    BIT  & $\checkmark$ & $\times$ & 88.93  & \textbf{82.34}  &  \textbf{67.38}\\
    BIT  & $\times$ & $\checkmark$ & 87.87  &  81.40  &  60.23\\
    BIT  &  $\checkmark$ & $\checkmark$ & \textbf{89.07} &  82.01  &  65.68\\
    \bottomrule
    \end{tabular}
    \label{tab:abalation_pe}
\end{table}
\begin{table}
    \centering
    \caption{Effect of the token length. The F1/IoU scores of the BIT are evaluated on the LEVIR-CD, WHU-CD, and DSIFN-CD test sets. Note that the depth of TE and TD are set to 1.}
    \begin{tabular}{c|cc|cc|cc}
    \toprule
     \multicolumn{1}{c}{} &
    \multicolumn{2}{c|}{\textbf{LEVIR-CD}}  &  \multicolumn{2}{c|}{\textbf{WHU-CD}}  &
    \multicolumn{2}{c}{\textbf{DSIFN-CD}}  \\
    Length  &  F1 & IoU &  F1 & IoU  & F1 & IoU  \\
    \midrule
    32 &  87.76 & 78.18 &  81.53 & 68.82 & 62.40 & 45.35\\
    16  &  88.45 & 79.74 &  81.79 & 69.19  & 63.07 & 46.06\\
    8  & 88.19 & 78.88  & 81.83 & 69.27  & 64.28 & 47.36\\
    4 & \textbf{88.93} & \textbf{80.07} &  \textbf{82.34} & \textbf{70.00}  & \textbf{67.38} & \textbf{50.80} \\
    2 & 88.90 & 80.02  & 82.02 & 69.53  & 65.13 & 48.29 \\
    \bottomrule
    \end{tabular}
    \label{tab:token_length}
\end{table}

\begin{table}
    \centering
    \caption{Effect of the depth of the transformer. We perform analysis on the Encoder Depth (E.D.) and Decoder Depth (D.D.) of the BIT, and report the F1/IoU scores for each configuration on the LEVIR-CD, WHU-CD, and DSIFN-CD test sets.}
    \resizebox{0.5\textwidth}{!}{
    \begin{tabular}{cc|cc|cc|cc}
    \toprule
     \multicolumn{2}{c}{} &
    \multicolumn{2}{c|}{\textbf{LEVIR-CD}}  &  \multicolumn{2}{c|}{\textbf{WHU-CD}}  &
    \multicolumn{2}{c}{\textbf{DSIFN-CD}}  \\
    E.D. &  D.D. & F1 & IoU & F1 & IoU & F1 & IoU\\
    \midrule
    1 & 1 & 88.93&80.07 & 82.34&70.00  & 67.38&50.80\\
    \midrule
    2 & 1 & 89.13&80.39 & 81.83&69.24 &66.96&50.34 \\
    4 & 1 & 88.97&80.13 & 82.15&69.70 & 66.95&50.32 \\
    8 & 1 & 88.93&80.06 & 80.73&67.68 & 67.11&50.50 \\
    \midrule
    1 & 2 & 88.91&80.03 & 82.99&70.92 & 67.17&50.57\\
    1 & 4 & 89.26&80.59 & 83.69&71.95 &69.05&52.73 \\
    1 & 8 & \textbf{89.31}& \textbf{80.68} & \textbf{83.98}&\textbf{72.39} & \textbf{69.26} & \textbf{52.97} \\
    \bottomrule
    \end{tabular}
    }
    \label{tab:transformer_depth}
\end{table}

\textbf{Training visualization}. Fig. \ref{fig:training_acc} illustrates the mean F1-score on the training/validation sets for each training epoch. We can observe that although the Base and BIT models have similar performance on training accuracy, BIT outperforms Base with regard to the validation accuracy in terms of stability and effectiveness. It indicates that the training of BIT is more stable and efficient, and our BIT-based model has more generalization ability. It may due to its ability to learn compact context-rich concepts, which effectively represent the change of interest. 

\subsection{Ablation studies}
\label{ssec:ablation}

\textbf{Context modeling}. We perform ablation on the Transformer Encoder (TE) to validate its effectiveness in context modeling, where multi-head self-attention is the core component in TE for modeling context. From Tab. \ref{tab:abalation_bit}, we can observe consistent and significant drops in F1-score on the LEVIR-CD, WHU-CD, and DSIFN-CD datasets when removing TE from BIT. It indicates the vital importance of self-attention in TE to model relations within token-based space-time. Moreover, we replace our BIT with a Non-local \cite{Wang2018b} self-attention layer, which is able to model relations within the pixel-based space-time. The comparison results in Tab. \ref{tab:abalation_bit} show our BIT outperforms Non-local on the three test sets with a significant margin. It may because our BIT learns the context in a tokens-based space, which is more compact and has higher information density than that of Non-local, thus facilitating the effective extraction of relations.

\textbf{Ablation on tokenizer}. We perform ablation on the tokenizer by removing it from the BIT. The resulting model can be considered to use dense tokens, which are sequences of features extracted by the CNN backbone. As shown in Tab. \ref{tab:abalation_bit}, the BIT-based model (w.o. tokenizer) receives significant drops in the F1-score. It indicates that the tokenizer module is critical in our transformer-based framework. We can see that the model (w.o. tokenizer) only slightly better than Base\_S4. It may because that the dense features contain too much redundancy information that makes the training of the transformer-based model a tough task. On the contrary, our proposed tokenizer spatially pool the dense features to aggregate the semantic information, thus obtaining compact tokens of concepts. 

\textbf{Ablation on transformer decoder}. To verify the effectiveness of our Transformer Decoder (TD), we replace it with a simple module to fuse the tokens $\mathbf{T}_{new}^{i}$ from TE and the original features $\mathbf{X}^{i}$ from the CNN backbone. In the simple module, we expand the spatial dimension of each token in $\mathbf{T}_{new}^{i}$ (containing $L$ tokens) to a shape of $\mathbb{R}^{HW}$. And the $L$ expanded tokens and $\mathbf{X}^{i}$ are summed to produce the updated features that are then fed to the prediction head. Tab. \ref{tab:abalation_bit} indicates consistent performance declines of the BIT model without TD on the three test sets. It may because cross-attention (the core part of TD) provides an elegant way to enhance the original features with the context-rich tokens by modeling their relations. Furthermore, the BIT (w.o. both TE and TD) is much inferior to the normal BIT model.

\textbf{Effect of position embedding}. The Transformer architecture is permutation-invariant, while the CD task requires both spatial and temporal position information. To this end, we add the learned position embedding (PE) to the feature sequence fed to the transformer. We perform ablations on PE in TE and TD. We set the BIT model containing no PE as the baseline. As shown in Tab. \ref{tab:abalation_pe}, our BIT model achieves consistent improvements in the F1-score on the three test sets when adding PE to the tokens fed into TE. It indicates that the position information within the bitemporal token sets is critical for context modeling in TE. Compared to the baseline, there are no significant improvements in the F1-score to the BIT model when adding PE to queries fed into TD. The positional information may be unnecessary to the queries into TD because the keys (i.e., tokens) into TD are highly abstract and contain no spatial structure. Therefore, we only add PE in TE, but not in TD in our BIT model. 

\begin{figure}
    \centering
    \includegraphics[width=0.5\textwidth]{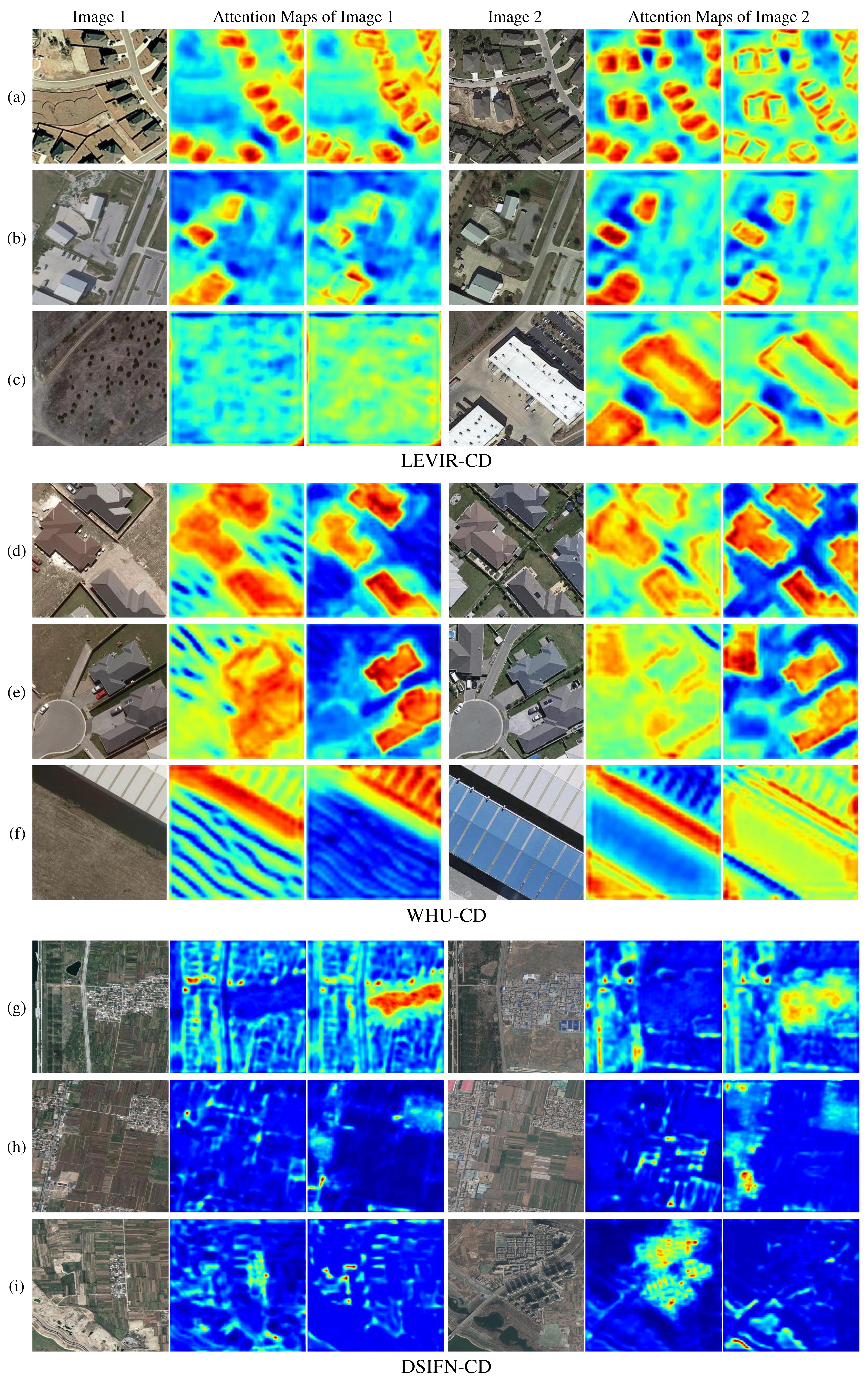}
    \caption{Token visualization on the LEVIR-CD, WHU-CD, and DSIFN-CD test sets. Red denotes higher attention values and blue denotes lower values.}
    \label{fig:visulize_tokens_all}
\end{figure}

\subsection{Parameter analysis} \label{parameter_analysis}

\textbf{Token length}. Our tokenizer spatially pools the dense features of the image into a compact token set. Our intuition is that the change of interest within the bitemporal images can be described by a few visual concepts, i.e., semantic tokens. The length of the token set $L$ is an important hyperparameter. We test different $L \in \{2,4,8,16,32\}$ to analyze its effect on the performance of our model on the LEVIR-CD, WHU-CD, and DSIFN-CD dataset, respectively. Tab. \ref{tab:token_length} shows a significant improvement in the F1-score of the model when reducing the token length from 32 to 4. It indicates that a compact token set is sufficient to denote semantic concepts of interest changes and redundant tokens may hinder the model performance. We can also observe a slight drop in F1-score when further decreasing $L$ from 4 to 2. It is because the model may lose some useful information related to change concepts when $L$ is too short. Therefore, we set $L$ to 4.

\textbf{Depth of transformer}. The number of transformer layers is one important hyperparameter. We test different configurations of the BIT model that contains varying numbers of transformer layers in TE and TD. Tab. \ref{tab:transformer_depth} shows no significant improvements to the F1/IoU scores of BIT on the three datasets when increasing the depth of the transformer encoder. It indicates that relations between the bitemporal tokens can be well learned by a single layer TE. Tab. \ref{tab:transformer_depth} also shows the model performance is roughly positively correlated with the decoder depth. It may because image features are refined after each layer of the transformer decoder by considering the context-rich tokens. The best result is obtained when the decoder depth is 8. Although there may be performance gains by further increasing the decoder depth, for the tradeoff between efficiency and precision, we set the encoder depth to 1 and the decoder depth to 8.

\begin{figure*}
    \centering
    \includegraphics[width=\textwidth]{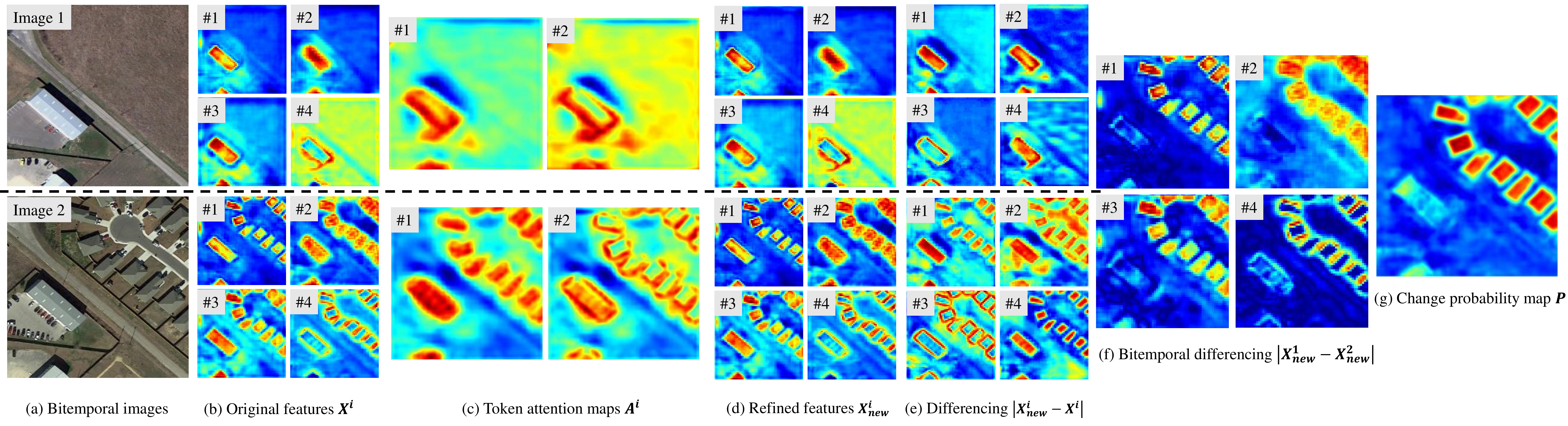}
    \caption{An example of network visualization. (a) input images, (b) selected high-level feature maps $\mathbf{X}^{i}$, (c) selected attention maps $\mathbf{A}^{i}$ by tokenizer, (d) refined feature maps $\mathbf{X}^{i}_{new}$, (e) differencing between $\mathbf{X}^{i}_{new}$ and $\mathbf{X}^{i}$, (f) bitemporal feature differencing image, (g) change probability map $\mathbf{P}$. The sample is from the LEVIR-CD data set. We use the same normalization (min-max) to visualize each activation map.}
    \label{fig:network_vis}
\end{figure*}

\subsection{Token visualization}

We hypothesize that our tokenizer can extract high-level semantic concepts that reveal the change of interest. For better understanding the semantic tokens, we visualize the attention maps $\mathbf{A}^{i} \in \mathbb{R}^{HW \times L}$ that the tokenizer extracted from the bitemporal feature maps. Each token $\mathbf{T}^{i}_{l}$ in the token set $\mathbf{T}^{i}$ is corresponding to one attention map $\mathbf{A}^{i}_{l} \in \mathbb{R}^{HW}$. Fig. \ref{fig:visulize_tokens_all} shows the visualization results of tokens for some bitemporal images from the LEVIR-CD, WHU-CD, and DSIFN-CD datasets. We display the attention maps of two selected tokens from $\mathbf{T}^{i}$ for each input image. Red denotes higher attention values and blue denotes lower values. 

From Fig. \ref{fig:visulize_tokens_all}, we can see that the extracted token can attend to the region that belongs to the semantic concept of the change of interest. Different tokens may relate to objects of different semantic meanings. For example, as the LEVIR-CD and WHU-CD datasets only describe the building changes, the learned tokens in these datasets mainly attend to the pixels belongs to buildings. While because the DSIFN-CD dataset contains various kinds of changes, these tokens can highlight different semantic areas, such as buildings, croplands, and water bodies. Interestingly, as shown in Fig. \ref{fig:visulize_tokens_all} (c) and (f), our tokenizer can also highlight the pixels surrounding the building (e.g., shadow), even though no explicit supervision of such areas is provided when training our model. It is not surprising because the context surrounding the building is a critical cue for object recognition. It indicates that our model can implicitly learn some additional concepts to promote change recognition.

\subsection{Network visualization}
To better understand our model, we provide an example to visualize the activation maps at different stages of the BIT model. Given the bitemporal image (Fig. \ref{fig:network_vis} (a)), a Siamese FCN generates the high-level feature maps $\mathbf{X}^{i}$ (Fig. \ref{fig:network_vis} (b)). Then the tokenizer spatially pools the feature maps into several token vectors using the learned attention maps $\mathbf{A}^{i}$ (Fig. \ref{fig:network_vis} (c)). The context-rich tokens generated by the transformer encoder are then projected back to the pixel-space via the transformer decoder, resulting in the refined feature maps $\mathbf{X}^{i}_{new}$ (Fig. \ref{fig:network_vis} (d)). We show four corresponding representative feature maps from the original features $\mathbf{X}^{i}$, and from the refined features $\mathbf{X}^{i}_{new}$. From Fig. \ref{fig:network_vis} (b) and (d), we can observe that our model can extract high-level features related to the change of interest for each temporal image, such as concepts of buildings and their edges. To better illustrate the effect of the BIT module, the differencing images between the refined and the original features are shown in Fig. \ref{fig:network_vis} (e). It indicates that our BIT can further highlight the regions of semantic concepts related to the change category. Lastly, the prediction head calculates feature differencing images (Fig. \ref{fig:network_vis} (f)) between $\mathbf{X}^{i}_{new}$ and $\mathbf{X}^{i}$, and generates the change probability map $\mathbf{P}$ (Fig. \ref{fig:network_vis} (g)).

\section{Discussion}
\label{sec:discussion}
We provide an efficient and effective method to perform change detection in high-resolution remote sensing images. The high reflectance variation for pixels of the same category in whole space-time brings difficulties to the model in recognizing objects of interest and distinguishing real changes from irrelevant changes. Context modeling in space-time is critical for enhancing feature discrimination power. Our proposed BIT module can efficiently model the context information in the token-based space-time and use the context-rich tokens to enhance the original features. Compared to the Base model, our BIT-base model can generate more accurate predictions with fewer false alarms and higher recalls (see Fig. \ref{fig:visualization_all} and Table \ref{tab:comparison_sotas}). Furthermore, the BIT can enhance the efficiency and stability of the training of the model (see Fig. \ref{fig:training_acc}). It is because that our BIT expresses the images into a small number of visual words (token vectors), such high-density information may improve the training efficiency. Our BIT can also be viewed as an efficient attention-base way to increase the reception field of the model, thus benefit feature representation power for change recognition.


\section{Conclusion}
\label{sec:conclusion}
In this paper, we propose an efficient transformer-based model for change detection in remote sensing images. Our BIT learns a compact set of tokens to represent high-level concepts that reveal the change of interest existing in the bitemporal images. We leverage the transformer to relate semantic concepts in the token-based space-time. Extensive experiments have validated the effectiveness of our method. We replace the last convolutional stage of ResNet18 with BIT, obtaining significant accuracy improvements (1.7/2.4/10.8 points of the F1-score on the LEVIR-CD/WHU-CD/DSIFN-CD test sets) with 3 times lower computational complexity and 3 times smaller model parameters. Our empirical evidence indicates BIT is more efficient and effective than purely convolutional modules. Only using a simple CNN backbone (ResNet18), our method outperforms several other CD methods that employ more sophisticated structures, such as FPN and UNet. We also show better performance in terms of efficiency and accuracy than four recent attention-based methods on the three CD datasets.





\ifCLASSOPTIONcaptionsoff
  \newpage
\fi

{\small
\bibliographystyle{IEEEtran}
\bibliography{refs}

\begin{thebibliography}{10}
\providecommand{\url}[1]{#1}
\csname url@samestyle\endcsname
\providecommand{\newblock}{\relax}
\providecommand{\bibinfo}[2]{#2}
\providecommand{\BIBentrySTDinterwordspacing}{\spaceskip=0pt\relax}
\providecommand{\BIBentryALTinterwordstretchfactor}{4}
\providecommand{\BIBentryALTinterwordspacing}{\spaceskip=\fontdimen2\font plus
\BIBentryALTinterwordstretchfactor\fontdimen3\font minus
  \fontdimen4\font\relax}
\providecommand{\BIBforeignlanguage}[2]{{%
\expandafter\ifx\csname l@#1\endcsname\relax
\typeout{** WARNING: IEEEtran.bst: No hyphenation pattern has been}%
\typeout{** loaded for the language `#1'. Using the pattern for}%
\typeout{** the default language instead.}%
\else
\language=\csname l@#1\endcsname
\fi
#2}}
\providecommand{\BIBdecl}{\relax}
\BIBdecl

\bibitem{SINGH1989}
A.~SINGH, ``Review article digital change detection techniques using
  remotely-sensed data,'' \emph{International Journal of Remote Sensing},
  vol.~10, no.~6, pp. 989--1003, 1989.

\bibitem{Chen2020e}
H.~Chen and Z.~Shi, ``A spatial-temporal attention-based method and a new
  dataset for remote sensing image change detection,'' \emph{Remote. Sens.},
  vol.~12, no.~10, p. 1662, 2020.

\bibitem{Bem2020}
P.~P. de~{Bem}, O.~A. de~Carvalho~{Junior}, R.~F. {Guimarães}, and R.~A.~T.
  {Gomes}, ``Change detection of deforestation in the brazilian amazon using
  landsat data and convolutional neural networks,'' \emph{Remote Sensing},
  vol.~12, no.~6, p. 901, 2020.

\bibitem{Xu2019}
J.~Z. Xu, W.~Lu, Z.~Li, P.~Khaitan, and V.~Zaytseva, ``Building damage
  detection in satellite imagery using convolutional neural networks,'' 2019.

\bibitem{Shi2020}
W.~Shi, M.~Zhang, R.~Zhang, S.~Chen, and Z.~Zhan, ``Change detection based on
  artificial intelligence: State-of-the-art and challenges,'' \emph{Remote
  Sensing}, vol.~12, p. 1688, 2020.

\bibitem{Chen2020}
J.~Chen, Z.~Yuan, J.~Peng, L.~Chen, H.~Huang, J.~Zhu, T.~Lin, and H.~Li,
  ``Dasnet: Dual attentive fully convolutional siamese networks for change
  detection of high resolution satellite images.''

\bibitem{Zhang2019c}
M.~Zhang, G.~Xu, K.~Chen, M.~Yan, and X.~Sun, ``Triplet-based semantic relation
  learning for aerial remote sensing image change detection,'' \emph{{IEEE}
  Geosci. Remote. Sens. Lett.}, vol.~16, no.~2, pp. 266--270, 2019.

\bibitem{Zhang2020a}
M.~Zhang and W.~Shi, ``A feature difference convolutional neural network-based
  change detection method,'' \emph{TGRS}, pp. 1--15, 2020.

\bibitem{Liu2019b}
Y.~{Liu}, C.~{Pang}, Z.~{Zhan}, X.~{Zhang}, and X.~{Yang}, ``Building change
  detection for remote sensing images using a dual-task constrained deep
  siamese convolutional network model,'' \emph{IEEE Geoscience and Remote
  Sensing Letters}, pp. 1--5, 2020.

\bibitem{Zhang2020b}
C.~Zhang, P.~Yue, D.~Tapete, L.~Jiang, B.~Shangguan, L.~Huang, and G.~Liu, ``A
  deeply supervised image fusion network for change detection in high
  resolution bi-temporal remote sensing images,'' \emph{ISPRS}, vol. 166, pp.
  183--200, 2020.

\bibitem{Peng2020a}
X.~Peng, R.~Zhong, Z.~Li, and Q.~Li, ``Optical remote sensing image change
  detection based on attention mechanism and image difference,'' \emph{{IEEE}
  Transactions on Geoscience and Remote Sensing}, pp. 1--12, 2020.

\bibitem{Jiang2020}
H.~Jiang, X.~Hu, K.~Li, J.~Zhang, J.~Gong, and M.~Zhang, ``Pga-siamnet: Pyramid
  feature-based attention-guided siamese network for remote sensing
  orthoimagery building change detection,'' \emph{Remote Sensing}, vol.~12,
  no.~3, p. 484, 2020.

\bibitem{Diakogiannis2020}
F.~I. Diakogiannis, F.~Waldner, and P.~Caccetta, ``Looking for change? roll the
  dice and demand attention.''

\bibitem{Fang2021}
S.~Fang, K.~Li, J.~Shao, and Z.~Li, ``Snunet-cd: A densely connected siamese
  network for change detection of vhr images,'' \emph{{IEEE} Geoscience and
  Remote Sensing Letters}, pp. 1--5, 2021.

\bibitem{Vaswani2017}
A.~Vaswani, N.~Shazeer, N.~Parmar, J.~Uszkoreit, L.~Jones, A.~N. Gomez,
  L.~Kaiser, and I.~Polosukhin, ``Attention is all you need,'' in
  \emph{Advances in Neural Information Processing Systems 30: Annual Conference
  on Neural Information Processing Systems 2017, December 4-9, 2017, Long
  Beach, CA, {USA}}, I.~Guyon, U.~von Luxburg, S.~Bengio, H.~M. Wallach,
  R.~Fergus, S.~V.~N. Vishwanathan, and R.~Garnett, Eds., 2017, pp. 5998--6008.

\bibitem{Nemoto2017}
K.~Nemoto, T.~Imaizumi, S.~Hikosaka, R.~Hamaguchi, M.~Sato, and A.~Fujita,
  ``Building change detection via a combination of cnns using only rgb aerial
  imageries,'' Oct. 2017.

\bibitem{Ji2019}
S.~Ji, Y.~Shen, M.~Lu, and Y.~Zhang, ``Building instance change detection from
  large-scale aerial images using convolutional neural networks and simulated
  samples,'' \emph{Remote Sensing}, vol.~11, no.~11, p. 1343, 2019.

\bibitem{Liu2019a}
R.~Liu, M.~Kuffer, and C.~Persello, ``The temporal dynamics of slums employing
  a cnn-based change detection approach,'' \emph{Remote. Sens.}, vol.~11,
  no.~23, p. 2844, 2019.

\bibitem{Daudt2018a}
R.~C. Daudt, B.~L. Saux, A.~Boulch, and Y.~Gousseau, ``Urban change detection
  for multispectral earth observation using convolutional neural networks,'' in
  \emph{IGARSS}, 2018.

\bibitem{Rahman2018}
F.~U. Rahman, B.~Vasu, J.~V. Cor, J.~Kerekes, and A.~E. Savakis, ``Siamese
  network with multi-level features for patch-based change detection in
  satellite imagery,'' in \emph{2018 {IEEE} Global Conference on Signal and
  Information Processing, GlobalSIP 2018, Anaheim, CA, USA, November 26-29,
  2018}.\hskip 1em plus 0.5em minus 0.4em\relax {IEEE}, 2018, pp. 958--962.

\bibitem{Wang2020}
M.~Wang, K.~Tan, X.~Jia, X.~Wang, and Y.~Chen, ``A deep siamese network with
  hybrid convolutional feature extraction module for change detection based on
  multi-sensor remote sensing images,'' \emph{Remote Sensing}, vol.~12, no.~2,
  p. 205, 2020.

\bibitem{Daudt2018}
R.~C. Daudt, B.~L. Saux, and A.~Boulch, ``Fully convolutional siamese networks
  for change detection,'' in \emph{ICIP}, 2018.

\bibitem{Lebedev2018}
M.~A. Lebedev, Y.~V. Vizilter, O.~V. Vygolov, V.~A. Knyaz, and A.~Y. Rubis,
  ``Change detection in remote sensing images using conditional adversarial
  networks,'' vol. XLII-2, 2018, pp. 565--571.

\bibitem{Peng2019}
D.~{Peng}, Y.~{Zhang}, and H.~{Guan}, ``End-to-end change detection for high
  resolution satellite images using improved unet++,'' \emph{Remote Sensing},
  vol.~11, no.~11, p. 1382, 2019.

\bibitem{Bao2020}
T.~Bao, C.~Fu, T.~Fang, and H.~Huo, ``Ppcnet: A combined patch-level and
  pixel-level end-to-end deep network for high-resolution remote sensing image
  change detection,'' vol.~PP, pp. 1--5, 2020.

\bibitem{Hou2020}
B.~{Hou}, Q.~{Liu}, H.~{Wang}, and Y.~{Wang}, ``From w-net to cdgan: Bitemporal
  change detection via deep learning techniques,'' \emph{IEEE Transactions on
  Geoscience and Remote Sensing}, vol.~58, no.~3, pp. 1790--1802, 2020.

\bibitem{Zhan2017}
Y.~Zhan, K.~Fu, M.~Yan, X.~Sun, H.~Wang, and X.~Qiu, ``Change detection based
  on deep siamese convolutional network for optical aerial images,'' \emph{IEEE
  Geoscience and Remote Sensing Letters}, vol.~14, pp. 1845--1849, 2017.

\bibitem{Fang2019}
B.~{Fang}, L.~{Pan}, and R.~{Kou}, ``Dual learning-based siamese framework for
  change detection using bi-temporal vhr optical remote sensing images,''
  \emph{Remote Sensing}, vol.~11, no.~11, p. 1292, 2019.

\bibitem{Zhao2020c}
W.~Zhao, X.~Chen, X.~Ge, and J.~Chen, ``Using adversarial network for multiple
  change detection in bitemporal remote sensing imagery,'' \emph{{IEEE}
  Geoscience and Remote Sensing Letters}, pp. 1--5, 2020.

\bibitem{Chen2021a}
H.~{Chen}, W.~{Li}, and Z.~{Shi}, ``Adversarial instance augmentation for
  building change detection in remote sensing images,'' \emph{IEEE Transactions
  on Geoscience and Remote Sensing}, pp. 1--16, 2021.

\bibitem{Zhao2020b}
W.~Zhao, L.~Mou, J.~Chen, Y.~Bo, and W.~J. Emery, ``Incorporating metric
  learning and adversarial network for seasonal invariant change detection,''
  \emph{{IEEE} Trans. Geosci. Remote. Sens.}, vol.~58, no.~4, pp. 2720--2731,
  2020.

\bibitem{He2016}
K.~He, X.~Zhang, S.~Ren, and J.~Sun, ``Deep residual learning for image
  recognition,'' in \emph{2016 {IEEE} Conference on Computer Vision and Pattern
  Recognition, {CVPR} 2016, Las Vegas, NV, USA, June 27-30, 2016}.\hskip 1em
  plus 0.5em minus 0.4em\relax {IEEE} Computer Society, 2016, pp. 770--778.

\bibitem{Dosovitskiy2020}
A.~Dosovitskiy, L.~Beyer, A.~Kolesnikov, D.~Weissenborn, X.~Zhai,
  T.~Unterthiner, M.~Dehghani, M.~Minderer, G.~Heigold, S.~Gelly, J.~Uszkoreit,
  and N.~Houlsby, ``An image is worth 16x16 words: Transformers for image
  recognition at scale.''

\bibitem{Touvron2020}
H.~Touvron, M.~Cord, M.~Douze, F.~Massa, A.~Sablayrolles, and H.~Jégou,
  ``Training data-efficient image transformers \& distillation through
  attention.''

\bibitem{Wu2020}
B.~Wu, C.~Xu, X.~Dai, A.~Wan, P.~Zhang, M.~Tomizuka, K.~Keutzer, and P.~Vajda,
  ``Visual transformers: Token-based image representation and processing for
  computer vision,'' \emph{CoRR}, vol. abs/2006.03677, 2020.

\bibitem{Zhang2020c}
D.~Zhang, H.~Zhang, J.~Tang, M.~Wang, X.~Hua, and Q.~Sun, ``Feature pyramid
  transformer,'' in \emph{Computer Vision -- ECCV 2020}, A.~Vedaldi,
  H.~Bischof, T.~Brox, and J.-M. Frahm, Eds.\hskip 1em plus 0.5em minus
  0.4em\relax Cham: Springer International Publishing, 2020, pp. 323--339.

\bibitem{Zheng2020}
S.~Zheng, J.~Lu, H.~Zhao, X.~Zhu, Z.~Luo, Y.~Wang, Y.~Fu, J.~Feng, T.~Xiang,
  P.~H.~S. Torr, and L.~Zhang, ``Rethinking semantic segmentation from a
  sequence-to-sequence perspective with transformers.''

\bibitem{Carion2020}
N.~Carion, F.~Massa, G.~Synnaeve, N.~Usunier, A.~Kirillov, and S.~Zagoruyko,
  ``End-to-end object detection with transformers,'' in \emph{Computer Vision -
  {ECCV} 2020 - 16th European Conference, Glasgow, UK, August 23-28, 2020,
  Proceedings, Part {I}}, ser. Lecture Notes in Computer Science, A.~Vedaldi,
  H.~Bischof, T.~Brox, and J.~Frahm, Eds., vol. 12346.\hskip 1em plus 0.5em
  minus 0.4em\relax Springer, 2020, pp. 213--229.

\bibitem{Zhu2021}
\BIBentryALTinterwordspacing
X.~Zhu, W.~Su, L.~Lu, B.~Li, X.~Wang, and J.~Dai, ``Deformable {\{}detr{\}}:
  Deformable transformers for end-to-end object detection,'' in
  \emph{International Conference on Learning Representations}, 2021. [Online].
  Available: \url{https://openreview.net/forum?id=gZ9hCDWe6ke}
\BIBentrySTDinterwordspacing

\bibitem{Chen2020h}
M.~Chen, A.~Radford, R.~Child, J.~Wu, H.~Jun, D.~Luan, and I.~Sutskever,
  ``Generative pretraining from pixels,'' in \emph{Proceedings of the 37th
  International Conference on Machine Learning, {ICML} 2020, 13-18 July 2020,
  Virtual Event}, ser. Proceedings of Machine Learning Research, vol.
  119.\hskip 1em plus 0.5em minus 0.4em\relax {PMLR}, 2020, pp. 1691--1703.

\bibitem{Esser2020}
P.~Esser, R.~Rombach, and B.~Ommer, ``Taming transformers for high-resolution
  image synthesis.''

\bibitem{Liu2021}
W.~Liu, S.~Chen, L.~Guo, X.~Zhu, and J.~Liu, ``Cptr: Full transformer network
  for image captioning.''

\bibitem{Yang2020c}
F.~Yang, H.~Yang, J.~Fu, H.~Lu, and B.~Guo, ``Learning texture transformer
  network for image super-resolution,'' jun 2020.

\bibitem{Chen2020g}
H.~Chen, Y.~Wang, T.~Guo, C.~Xu, Y.~Deng, Z.~Liu, S.~Ma, C.~Xu, C.~Xu, and
  W.~Gao, ``Pre-trained image processing transformer,'' 2020.

\bibitem{Yuan2020}
Y.~{Yuan} and L.~{Lin}, ``Self-supervised pre-training of transformers for
  satellite image time series classification,'' \emph{IEEE Journal of Selected
  Topics in Applied Earth Observations and Remote Sensing}, pp. 1--1, 2020.

\bibitem{Li2020c}
Z.~{Li}, G.~{Chen}, and T.~{Zhang}, ``A cnn-transformer hybrid approach for
  crop classification using multitemporal multisensor images,'' \emph{IEEE
  Journal of Selected Topics in Applied Earth Observations and Remote Sensing},
  vol.~13, pp. 847--858, 2020.

\bibitem{He2020a}
J.~{He}, L.~{Zhao}, H.~{Yang}, M.~{Zhang}, and W.~{Li}, ``Hsi-bert:
  Hyperspectral image classification using the bidirectional encoder
  representation from transformers,'' \emph{IEEE Transactions on Geoscience and
  Remote Sensing}, vol.~58, no.~1, pp. 165--178, 2020.

\bibitem{Bazi2021}
Y.~Bazi, L.~Bashmal, M.~M.~A. Rahhal, R.~A. Dayil, and N.~A. Ajlan, ``Vision
  transformers for remote sensing image classification,'' \emph{Remote
  Sensing}, vol.~13, no.~3, 2021.

\bibitem{Shen2020a}
X.~Shen, B.~Liu, Y.~Zhou, and J.~Zhao, ``Remote sensing image caption
  generation via transformer and reinforcement learning,'' \emph{Multim. Tools
  Appl.}, vol.~79, no. 35-36, pp. 26\,661--26\,682, 2020.

\bibitem{Wang2020e}
Q.~{Wang}, W.~{Huang}, X.~{Zhang}, and X.~{Li}, ``Word-sentence framework for
  remote sensing image captioning,'' \emph{IEEE Transactions on Geoscience and
  Remote Sensing}, pp. 1--12, 2020.

\bibitem{Devlin2019}
J.~Devlin, M.-W. Chang, K.~Lee, and K.~Toutanova, ``{BERT}: Pre-training of
  deep bidirectional transformers for language understanding,'' in
  \emph{Proceedings of the 2019 Conference of the North {A}merican Chapter of
  the Association for Computational Linguistics: Human Language Technologies,
  Volume 1 (Long and Short Papers)}.\hskip 1em plus 0.5em minus 0.4em\relax
  Minneapolis, Minnesota: Association for Computational Linguistics, Jun. 2019,
  pp. 4171--4186.

\bibitem{Nguyen2019}
T.~Q. Nguyen and J.~Salazar, ``Transformers without tears: Improving the
  normalization of self-attention,'' \emph{CoRR}, vol. abs/1910.05895, 2019.

\bibitem{Hendrycks2016}
D.~Hendrycks and K.~Gimpel, ``Gaussian error linear units (gelus).''

\bibitem{Ji2019a}
S.~Ji, S.~Wei, and M.~Lu, ``Fully convolutional networks for multisource
  building extraction from an open aerial and satellite imagery data set,''
  \emph{{IEEE} Trans. Geoscience and Remote Sensing}, vol.~57, no.~1, pp.
  574--586, 2019.

\bibitem{Zhou2018}
\BIBentryALTinterwordspacing
Z.~Zhou, M.~M.~R. Siddiquee, N.~Tajbakhsh, and J.~Liang, ``Unet++: {A} nested
  u-net architecture for medical image segmentation,'' in \emph{Deep Learning
  in Medical Image Analysis - and - Multimodal Learning for Clinical Decision
  Support - 4th International Workshop, {DLMIA} 2018, and 8th International
  Workshop, {ML-CDS} 2018, Held in Conjunction with {MICCAI} 2018, Granada,
  Spain, September 20, 2018, Proceedings}, ser. Lecture Notes in Computer
  Science, vol. 11045.\hskip 1em plus 0.5em minus 0.4em\relax Springer, 2018,
  pp. 3--11. [Online]. Available:
  \url{https://doi.org/10.1007/978-3-030-00889-5_1}
\BIBentrySTDinterwordspacing

\bibitem{Wang2018b}
X.~Wang, R.~Girshick, A.~Gupta, and K.~He, ``Non-local neural networks,'' jun
  2018.

\end{thebibliography}
}

\end{document}